\PassOptionsToPackage{table}{xcolor}
\ifx\iflatexml\undefined
  \IfFileExists{google.cls}{%
    \documentclass[11pt, a4paper, twocolumn, copyright, goog]{google}%
    \def\UsingGoogleClass{1}%
  }{%
    \documentclass[11pt, a4paper, twocolumn]{article}%
  }
\else
  \documentclass[11pt, a4paper, twocolumn]{article}%
\fi
\usepackage{latexml} % \iflatexml: true only when arXiv builds the HTML version (LaTeXML); a local stand-in otherwise
\ifdefined\UsingGoogleClass\else% Local stand-in for the Overleaf "Google LaTeX Template" class (google.cls), used ONLY when google.cls is
\iflatexml\else
\usepackage[a4paper,top=2.3cm,bottom=2.4cm,left=1.9cm,right=1.9cm,columnsep=0.75cm,headheight=28pt,headsep=14pt]{geometry}
\usepackage[T1]{fontenc}
\usepackage{XCharter}
\usepackage{environ}
\usepackage{fancyhdr}
\usepackage{titlesec}
\usepackage{xcolor}
\titleformat{\section}{\large\bfseries}{\thesection.}{0.6em}{}
\titleformat{\subsection}{\normalsize\bfseries}{\thesubsection.}{0.6em}{}
\fi

\iflatexml
\providecommand{\affil}[2][]{}
\makeatletter
\else
\usepackage{authblk}

\makeatletter
\renewcommand\AB@affilsepx{, \protect\Affilfont}
\fi
\def\@gkeywords{}\def\@gcorr{}\def\@greport{}\def\@gurl{}\newif\if@glogo
\newcommand{\keywords}[1]{\gdef\@gkeywords{#1}}
\newcommand{\paperurl}[1]{\gdef\@gurl{#1}}
\newcommand{\uselogo}[1]{\@glogotrue}
\newcommand{\correspondingauthor}[1]{\gdef\@gcorr{#1}}
\newcommand{\reportnumber}[1]{\gdef\@greport{#1}}
\iflatexml\else
\global\let\@gstoredabstract\@empty
\RenewEnviron{abstract}{\global\let\@gstoredabstract\BODY}
\def\@glogo{{\sffamily\bfseries\Large\textcolor[HTML]{4285F4}{G}\textcolor[HTML]{EA4335}{o}\textcolor[HTML]{FBBC05}{o}\textcolor[HTML]{4285F4}{g}\textcolor[HTML]{34A853}{l}\textcolor[HTML]{EA4335}{e}}}
\fancypagestyle{gfirst}{\fancyhf{}%
  \fancyhead[L]{\if@glogo\@glogo\fi}%
  \fancyhead[R]{\scriptsize\itshape\ifx\@gurl\@empty\else\textcolor[HTML]{1F5A94}{\@gurl}\\\fi\@date}%
  \fancyfoot[L]{\scriptsize\ifx\@gcorr\@empty\else\textit{Corresponding author(s): \@gcorr}\\\fi\copyright\ \the\year\ Google. All rights reserved}%
  \fancyfoot[R]{\scriptsize Google Technical Report \@greport}%
  }
\fancypagestyle{grest}{\fancyhf{}\fancyhead[L]{\scriptsize\@title}\fancyfoot[C]{\small\thepage}%
  }
\renewcommand{\maketitle}{%
  \twocolumn[{%
    \thispagestyle{gfirst}%
    {\raggedright{\LARGE\bfseries\@title\par}\vspace{0.8em}{\small\bfseries\Authfont\AB@authlist\par}\vspace{0.2em}{\scriptsize\AB@affillist\par}}%
    \vspace{1.1em}%
    \ifx\@gstoredabstract\@empty\else{\small\bfseries\@gstoredabstract\par}\fi
    \ifx\@gkeywords\@empty\else\vspace{0.8em}{\small\textit{Keywords: \@gkeywords}\par}\fi
    \vspace{1.4em}}]}
\fi
\makeatother
\fi

\usepackage[authoryear, sort&compress, round]{natbib}
\usepackage{amsmath,amssymb}
\usepackage{graphicx}
\usepackage{caption}
\usepackage{booktabs}
\usepackage{array}
\usepackage{enumitem}
\usepackage{xcolor}
\usepackage{microtype}
\usepackage{etoolbox}
\usepackage{placeins}  % no [section] option: it would flush Figure 1 before the Introduction
\usepackage{adjustbox} % appendix: shrink an over-wide table to the text width
\iflatexml\else\usepackage{titletoc}\fi  % appendix table of contents (PDF only; LaTeXML has no titletoc)
\usepackage{xurl}      % break long URLs in the bibliography
\usepackage{hyperref}
\definecolor{LinkBlue}{HTML}{1F5A94}
\definecolor{ForestGreen}{rgb}{0,0,0} % abstract highlight shown in black in the Google version
\hypersetup{colorlinks=true,linkcolor=LinkBlue,citecolor=LinkBlue,urlcolor=LinkBlue}
\AtBeginDocument{\hypersetup{%
  pdftitle={DeLoop: Repeated Actions as Self-Supervision for Long-Horizon Agents},
  pdfauthor={Tianle Chen, Pushkar Singh, Ketan Todi, Biswajit Paria, Qijia He, Wenqi Wang, Yuwen Tan, Jike Zhong, Sugato Basu, Deepti Ghadiyaram, Arjun Akula}}}
\usepackage[capitalize,noabbrev]{cleveref}

\definecolor{GoodGreen}{HTML}{1E8449}
\definecolor{BadRed}{HTML}{D62828}
\newcommand{\good}[1]{{\boldmath\textbf{#1}}}
\newcommand{\bad}[1]{{\boldmath\textbf{#1}}}

\AtBeginEnvironment{table}{\let\good\goodcolored\let\bad\badcolored}
\AtBeginEnvironment{coltable}{\let\good\goodcolored\let\bad\badcolored}
\newcommand{\impr}[1]{\,{\scriptsize(#1)}}
\newcommand{\decl}[1]{\,{\scriptsize(#1)}}
\usepackage{xcolor}

\definecolor{TableHead}{HTML}{EEF2F5}
\definecolor{TableGroup}{HTML}{F6F7F8}
\definecolor{TableBase}{HTML}{FFF3D9}
\definecolor{TableOurs}{HTML}{E7F2F8}
\definecolor{TableCaution}{HTML}{FBE9E7}
\makeatletter
\renewcommand\table{\@dblfloat{table}}
\renewcommand\endtable{\end@dblfloat}
\makeatother
\makeatletter\newenvironment{coltable}[1][t]{\@float{table}[#1]}{\end@float}\makeatother
\iflatexml\else
\usepackage{environ}
\makeatletter
\newsavebox\g@tabbox
\let\g@origtabular\tabular \let\g@endorigtabular\endtabular
\renewenvironment{tabular}{\begin{lrbox}{\g@tabbox}\g@origtabular}%
  {\g@endorigtabular\end{lrbox}\ifdim\wd\g@tabbox>\linewidth\resizebox{\linewidth}{!}{\usebox\g@tabbox}\else\usebox\g@tabbox\fi}
\expandafter\let\expandafter\g@origtabstar\csname tabular*\endcsname
\expandafter\let\expandafter\g@endorigtabstar\csname endtabular*\endcsname
\RenewEnviron{tabular*}[2]{%
  \setbox\g@tabbox\hbox{\g@origtabular{#2}\BODY\g@endorigtabular}%
  \ifdim\wd\g@tabbox>#1\relax\resizebox{#1}{!}{\usebox\g@tabbox}\else\g@origtabstar{#1}{#2}\BODY\g@endorigtabstar\fi}
\makeatother
\fi

\uselogo{}

\title{DeLoop: Repeated Actions as Self-Supervision for Long-Horizon Agents}

\reportnumber{}

\iflatexml
\author{Tianle Chen$^{1,2}$ \and Pushkar Singh$^{1,\dagger}$ \and Ketan Todi$^{1}$ \and Biswajit Paria$^{1}$ \and Qijia He$^{1,3}$ \and Wenqi Wang$^{1,2}$ \and Yuwen Tan$^{1,2}$ \and Jike Zhong$^{4}$ \and Sugato Basu$^{1}$ \and Deepti Ghadiyaram$^{2}$ \and Arjun Akula$^{5,\dagger}$}
\else
\author[1,2]{Tianle Chen}
\author[1,$\dagger$]{Pushkar Singh}
\author[1]{Ketan Todi}
\author[1]{Biswajit Paria}
\author[1,3]{Qijia He}
\author[1,2]{Wenqi Wang}
\author[1,2]{Yuwen Tan}
\author[4]{Jike Zhong}
\author[1]{Sugato Basu}
\author[2]{Deepti Ghadiyaram}
\author[5,$\dagger$]{Arjun Akula}
\affil[1]{Google}
\affil[2]{Boston University}
\affil[3]{University of Washington}
\affil[4]{University of Southern California}
\affil[5]{Google DeepMind}
\affil[$\dagger$]{Project lead}
\fi

\iflatexml\else
\begin{abstract}

Language-model agents that run long tasks often redo work they have already done: they re-issue a tool call they made earlier and get back exactly the same observation, a step that adds no new information.
We identify this failure as a \emph{loop}.
Loops appear natively in Qwen, Llama, and Gemini agents and become far more common once the agent's history is compressed to fit its context window.
To reduce these loops, we propose \textbf{DeLoop}, a self-supervised training method.
DeLoop builds on a simple observation: the agent's own rollout can provide the supervision needed to learn from its loops.
We use a looping action as a negative example and the first later non-looping action in the same rollout as a positive example.
DeLoop learns from these pairs through direct preference optimization and does not require task rewards, gold answers, or LLM judges.
On CodeGym under compression, DeLoop reduces the share of Qwen3-8B tasks containing loops from 57.3\% to \textcolor{ForestGreen}{\textbf{6.1\%}}.
It also uses about one third as many steps per task, with similar task success.
On Qwen3-32B, \textbf{DeLoop} reduces loops and steps while raising task success from 37.7\% to 44.8\%. Code and data will be publicly available.

\end{abstract}
\fi

\begin{document}
% No page limit: let columns end where their text ends instead of stretching the space around headings.
\raggedbottom
\maketitle
\iflatexml
\noindent{\small $^{1}$Google, $^{2}$Boston University, $^{3}$University of Washington, $^{4}$University of Southern California, $^{5}$Google DeepMind, $^{\dagger}$Project lead}\par
\begin{abstract}

\end{abstract}
\makeatletter\noindent\textit{Keywords:} \@gkeywords\par\makeatother
\fi

\begin{figure*}[t]
\centering
\includegraphics[width=\textwidth]{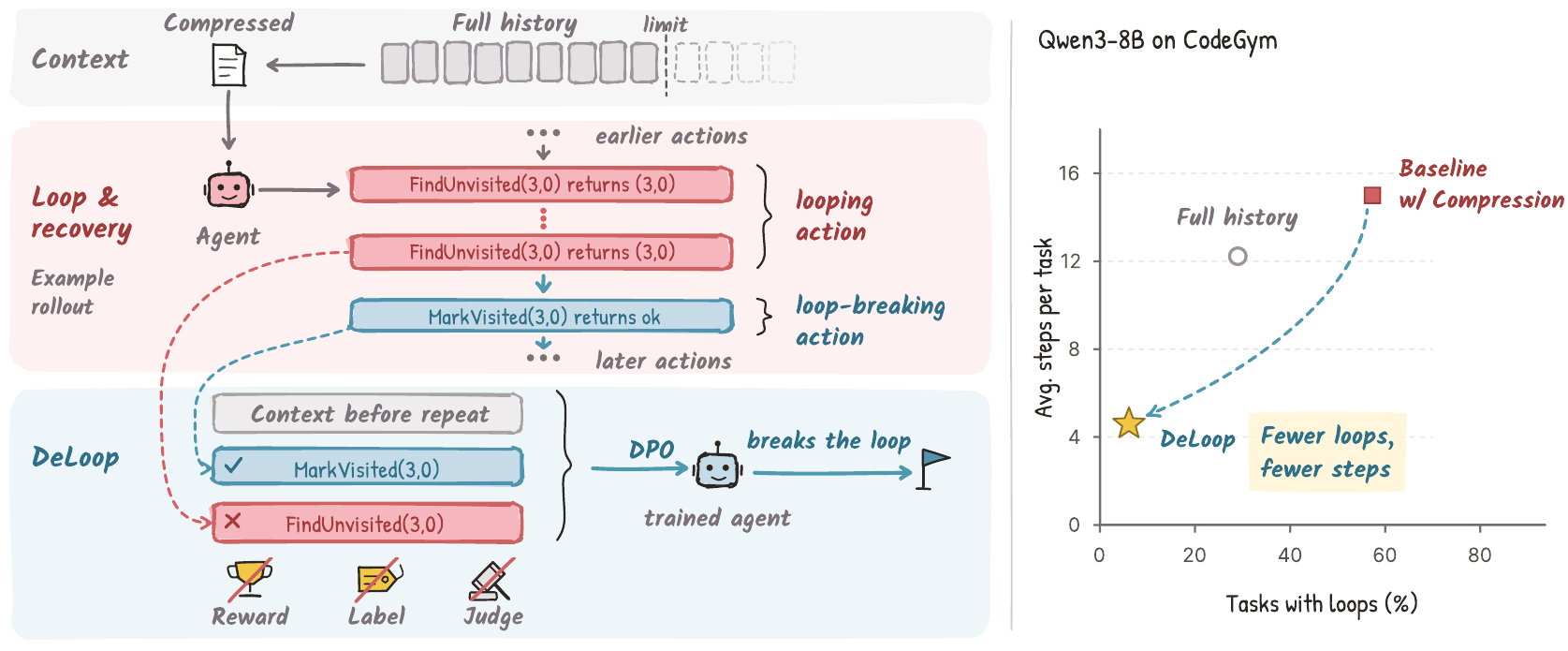}
\caption{\textbf{Training an agent on its own loops breaks the loop.}
\textit{(Left)} A repeated search and the later action that marks the server as visited form a preference pair. DPO trains the agent to prefer the recovery.
\textit{(Right)} Qwen3-8B on CodeGym: DeLoop and the baseline summarize at 4,096 tokens; full history is the uncompressed reference.}
\label{fig:main}\label{fig:miner}
\end{figure*}

% NAACL/ARR introduction (2026-09-28 prose pass, same content as 2026-09-09). Derived from 01_intro.tex; numbers via macros only.
\section{Introduction}
\label{sec:intro}

Language-model agents increasingly solve tasks over long interaction
horizons: they inspect repositories, execute commands, call tools, and act on
environment feedback across many steps
\citep{react,swe_agent,swe_bench,du2025generalizable}.
Yet they often redo work they have already done. The agent calls the same tool with the same arguments, receives the same observation, and then issues the same call again (\cref{fig:miner}). Each repeated call is a \emph{looping action}: it gets back an observation the agent already has and adds no new information. We formulate this failure as a \textbf{loop}.

The failure is already present even when the full, uncompressed history is available, but compression makes it worse. On CodeGym~\citep{du2025generalizable}, the loop rate of Qwen3-8B~\citep{qwen3}, the share of tasks with at least one loop, rises from $29.0\%$ with the full history (no summarization) to \bad{57.3\%} under summarization, while looping actions per task increase \bad{more than 3.5$\times$}!  
More importantly, in addition to wasted tool calls, we empirically observed that trajectories that enter a loop fail in all six Qwen3 and Gemini models we evaluate: on average, \bad{59.5\%}  of looping runs fail, compared with 33.4\% of loop-free runs (\cref{fig:loops-hurt}). 
%nearly one in four of its actions is a looping action.
% \deepti{remove my comment - -add directly to text}
% \deepti{an avg failure rate to make this more concrete} \sout{more often} 
Thus, context compression
makes looping more likely and increases the chances of task failure.
%amplifies a failure that is both costly and tied to failed tasks. 
% \deepti{this sentence is unclear -- is the failure costly? Or the amplification costly -- try to phrase in a simpler way}
%Compression thus makes looping a common, costly behavior that is tied to task failure as well.

Recent work manages growing interaction histories through external memory
\citep{memgpt,prolong}, hierarchical or summarized working context
\citep{hiagent,resum,contextfolding}, compact learned state
\citep{mem1}, and learned context compression or management
\citep{acon,supo,contextastool}.
These works study \emph{what} history should an agent retain to continue acting well under a bounded context window. In contrast, we study a complementary question about how the agent's policy uses its history after compression. That is, \emph{when the history records an action, does the agent use that record to avoid repeating it?} We find that most agents do so only partly. 

Another way to reduce looping is to improve the agent's policy. Some methods address agent failures at inference time through reflection~\citep{reflexion} or early stopping~\citep{runaway}. Others train agents using demonstrations~\citep{agenttuning,fireact}, search-based training data~\citep{agentr}, or step-level supervision~\citep{ipr2024,agenticdpo,toolgraph}. We instead ask: \textit{can the failure itself provide a useful signal to improve agent's policy?}

Based on these observations, we propose \textbf{DeLoop}, a self-supervised training method. Our key observation is that loops can provide their own training signal. We note that a loop is a simple trajectory-level event that can be identified directly from the action-observation trace, without requiring post-hoc failure attribution~\citep{agentrx,infiniteagentloops}. Within a single rollout, we pair a looping action with the \textbf{first later non-looping }action that differs from it. The looping action provides a natural negative, while the later action provides a positive signal. Figure~\ref{fig:main} illustrates this. These pairs are mined directly from the interaction, without gold answers, LLM judges, external rewards, or task-success labels. We train the policy on these pairs through direct preference optimization (DPO~\citealp{dpo}).

On Qwen3-8B, DeLoop cuts the loop rate \good{9.4$\times$} while lowering interaction cost, with similar task success. On Qwen3-32B, the self-mined pairs alone reduce loop rate from $38.5\%$ to \good{12.9\%} and raise task success from $37.7\%$ to \good{44.8\%}, while also lowering interaction cost.
Adding cross-domain pairs (DeLoop-Mix) raises Qwen3-8B task success from $25.1\%$ to \good{30.3\%}.
%\deepti{cross-mined is not explained so i'd recommend skipping it}
%\tianle{We have some related result about this in the main table, but I think it's ok to skip as the ablation of the mixture is not very completed}
Our contributions are:
\begin{enumerate}[leftmargin=1.4em,itemsep=1pt,topsep=2pt]
  \item \textbf{Identifying loops in long-horizon agents:} we identify and characterize loops, a simple failure in which an agent repeats an earlier action and receives the same observation. We show that loops occur even with the full history, grow sharply under compression, and are associated with task failure.

    \item \textbf{Mining loop actions as self-supervision:} we introduce a simple mining
    method that turns discovered loops into preference pairs from the agent's own
    rollouts, with no external supervision.
% \wenqi{Maybe we can say it another way? \textit{without gold answers, LLM judges, or task-success labels} repeats 3 times in introduction} 
\item \textbf{Reducing loops in long-horizon agents:} DeLoop reduces loops (57.3\% to \good{6.1\%} on Qwen3-8B) while lowering interaction cost. The gains also extend to larger models, where there is pronounced improvement in task success (45.1\% to \good{52.3\%} on Qwen3-14B and 37.7\% to \good{44.8\%} on Qwen3-32B).
%   \item \textbf{Reducing loops in long-horizon agents:} DeLoop reduces loops
% while lowering interaction cost. The gains also extend to larger models, where there is pronounced improvement in task success \deepti{add numbers} 
\end{enumerate}

% \wenqi{The introduction is clear! But the wording is somewhat repetitive, e.g. too many \textit{`much more often'}} 

% NAACL/ARR related work, final refinement of variant v1 (2026-10-12). Replaces sections/naacl_06_related.tex.
% New keys (in final.bib, each verified online 2026-10-12): agentbench2024, mast2025, ragen2025,
% stepdpo2024, agentq2024, score2025, her2017, ctrl2019.
% Dropped: productivestatespace (anonymous ARR submission, not verifiable), star2022 (cap of 8 new keys).
% Verifier pass 2026-10-12: all 8 new entries re-checked online; descriptions checked against abstracts/paper text (AgentBench v3 App. J.2.4, MAST FM-1.3, RAGEN Echo Trap, RLTR R_repeat, OpenHands stuck.py 4 consecutive pairs, TRACE abstract, ToolGraph/Agentic-DPO abstracts).
% TEMP 2026-10-12: new version (relwork workflow final); compare with naacl_06_related.tex and keep one.
\section{Related work}
\label{sec:related}

% DeLoop builds each preference pair from one rollout of the agent: an
% exact-repeat rule marks the looping action as rejected, and the agent's own
% later loop-breaking action is chosen, with no task outcome, gold action, or
% LLM judge.

\paragraph{Repeated actions in agents.}
Repetition is a known failure of LLM agents. AgentBench found that agents
mainly exceed their task limit because they repeat earlier
content~\citep{agentbench2024}. MAST also lists step repetition as a failure
mode of multi-agent systems~\citep{mast2025}, and RAGEN shows that multi-turn
RL can trap agents in repetitive reasoning~\citep{ragen2025}. Closest to our
setting, \citet{trace2026} find that repeated context compression increases
repeated exploration and improve the compression prompt to reduce it; we keep
the compressor fixed and change the policy. IAL-Scan finds code paths that can
run without bound~\citep{infiniteagentloops}, and other work locates agent
failures in execution traces~\citep{trail2025,agentrx,oat2026,whoandwhenpro2026,agentracer2025}.
At inference time, agents can reflect, stop early, or escalate to a stronger
model~\citep{reflexion,runaway,archer}. OpenHands flags an agent as ``stuck''
when the same action returns the same observation four times in a
row~\citep{openhands}. These methods do not change the agent policy. Even with a
blocker that refuses repeated calls and returns their earlier result, Qwen3-8B
on CodeGym still proposes an already executed call in 71.3\% of tasks, as often
as without the blocker (69.8\%); DeLoop lowers this to 13.6\%
(Appendix~\ref{app:checks:blocker}). We define a loop as a repeated tool call
that returns the same observation. Loops persist even with summaries designed
to keep past actions (Section~\ref{sec:compression}), so we train the agent
policy itself to break the loop.

\paragraph{Context management.}
Long-horizon agents need to fit their history into a limited context. Common context management methods
store the history in external memory or searchable logs~\citep{memgpt,prolong,memexrl},
summarize older interactions~\citep{hiagent,resum,contextfolding,ufold},
keep a compact learned state~\citep{mem1}, learn what to
keep~\citep{acon,supo,contextastool}, or validate their
summaries~\citep{mage}. However, removing plans from the context sharply hurts
performance~\citep{plansdontpersist}. These methods decide \emph{what}
history to keep. We instead ask whether the agent uses the kept history to avoid repeating actions.

\paragraph{Learning from agent trajectories.}
Most methods that train agents on trajectories need an external signal to
pick the better action. Agent tuning uses filtered or teacher-generated
trajectories~\citep{agenttuning,fireact,agentrefine}, or successful attempts and
outcome feedback~\citep{star2022,actre,seagent,multiharness2026,wang2024learningfromfailure,coevolving2025}. ETO and
Agentic-DPO take the better action from experts~\citep{song2024eto,agenticdpo}.
Step-DPO, IPR, and ToolGraph use step-level labels or
rewards~\citep{stepdpo2024,ipr2024,toolgraph}. Agent-R and Agent Q use tree
search guided by rewards or critiques~\citep{agentr,agentq2024}, and SCoRe
trains self-correction with online RL~\citep{score2025}. DeLoop instead
takes the better action from the agent's own later recovery in the same
rollout and trains with DPO~\citep{dpo}. Its only label is the exact-repeat
rule of Section~\ref{sec:definition}, so it needs no rewards, expert actions,
or step labels. Like hindsight experience replay~\citep{her2017}, it also
learns from failed rollouts.

\paragraph{Penalizing repetition.}
Repetition can also be penalized directly. RLTR adds a penalty on redundant
tool calls to its reward~\citep{rltr2025}. In text generation, penalized
sampling~\citep{ctrl2019}, unlikelihood training~\citep{welleck2020unlikelihood},
and DITTO~\citep{xu2022breakloop} reduce repeated tokens and sentences. These
penalties mark what to avoid; DeLoop also supplies the agent's own alternative
action. Using our loop flag as a per-step penalty in online RL (GRPO) lowers the
Qwen3-8B loop rate on CodeGym from 57.3\% to 15.0\%, against 6.1\% for DeLoop
(Section~\ref{sec:objectives}).

% plain_02_analysis.tex -- SPINE_v2 §2 row 3 (2026-10-08 merge of drafts/v2_analysis_fable_02_analysis.tex and
% drafts/v2_analysis_opus_02_analysis.tex). Section 2 "Action-State Collapse", placed before Related Work.
% Every result number enters via a \g macro (macros_gold.tex). Literal numbers left in the prose are compression
% settings (8,192 / 4,096 / 2,048-token budgets), not results. Detector equations verbatim from plain_03_method.tex.
% Loop example: verbatim from slide 6; the intermediate-turn count is omitted (GOLD_NUMBERS discrepancy 16).
% Failure / cost analyses: dataset not named in text (the slide 8/9 sources do not state it); only Gemini 3.6 Flash
% Gemini naming: AUTHOR DECISION 2026-10-08 -- the Flash model is Gemini 3.1 Flash.
% the source scripts (GOLD_NUMBERS #12; author check).
% Labels defined: sec:analysis, sec:definition, sec:loops-hurt, sec:compression.
% Labels referenced: fig:loops-hurt, fig:loop-cost (appendix), fig:compression, tab:battery (appendix),
%   app:setup, app:trueloop, sec:method.
\section{Loops in Long-Horizon Agents}
\label{sec:analysis}

\subsection{Defining a Loop}
\label{sec:definition}\label{sec:phenomenon}
A loop occurs when the agent repeats a previous tool call with identical
arguments and receives the same observation. Thus, that step adds no new
information.

For example, consider an agent counting clusters of online servers on a grid.
The agent has tools to find unvisited servers and mark them as visited.
In one rollout, it issues the same search three times without marking the
returned server. Because that server remains unmarked, each call returns
the same coordinates. The latter two calls therefore add no new information.
This rollout comes from a Qwen3-8B agent on CodeGym~\citep{du2025generalizable},
where agents solve tasks through sequences of tool calls.
Appendix~\ref{app:real-loop} shows the rollout excerpt.

We detect such repetitions by comparing each tool call and its observation
with earlier steps.
Let $a_t$ denote the tool call at step $t$, and $o_t$ its returned observation.
We denote a trajectory of $T$ steps by $h_T$.
With $\kappa$ canonicalizing each tool call so that
formatting differences do not count,
% \begin{gather*}
% s_t=\mathrm{1}\!\left[\exists i<t:\kappa(a_t)=\kappa(a_i)\ \wedge\ o_t=o_i\right],\\
% \operatorname{Loop}(h_T)=\mathrm{1}\!\left[\max_{t\leq T}s_t=1\right].
% \end{gather*}
% \biswa{Alternative definition:
let $x_t = \bigl(\kappa(a_t),\, o_t\bigr)$ denote the signature of step $t$.
Step $t$ is a \emph{repeat} if its signature appeared earlier:
\[
  s_t =
  \begin{cases}
    1 & \text{if } x_t = x_i \text{ for some } i < t, \\
    0 & \text{otherwise.}
  \end{cases}
\]
A trajectory $h_T$ \emph{loops} if it contains at least one repeat:
\[
  h_T \text{ loops} \iff s_t = 1 \text{ for some } t \le T.
\]

% }
A step with $s_t=1$ is a \emph{looping action}.
A repeated call with a new observation is not a looping action.
We then label any run containing at least one looping action as a
\emph{looping run}.
We report the percentage of looping runs as the \emph{loop rate}.

\subsection{Loops Are Costly}
\label{sec:loops-hurt}
Each looping action consumes tokens and a tool execution without yielding
a new observation. Under a fixed interaction budget, these repeated calls leave fewer
opportunities to take actions that could advance the task.
We compare looping and loop-free Qwen3-8B runs on CodeGym under
summarization at 4,096 tokens. Both groups include successful and failed runs.
Looping runs contain \bad{6.5} looping actions on average.
They execute \bad{10.5} more actions per task on average than loop-free runs.
Their interaction transcripts also contain
\bad{11k} more tokens.
The same pattern holds for all six models in the cost comparison
(Appendix Figure~\ref{fig:loop-cost}).

\begin{figure}[t]
\centering
\includegraphics[width=\linewidth]{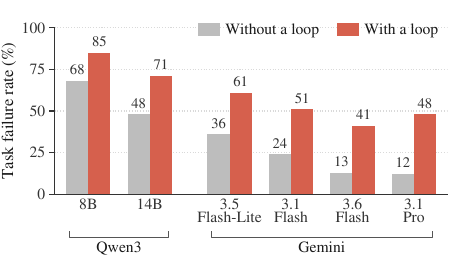}
\vspace{-8mm}
\caption{\textbf{Failure rates of looping and loop-free runs.}
Task-failure rate for runs with a loop (red) and without one (gray), for
Qwen3-8B, Qwen3-14B, and four Gemini models in their compressed settings.
Detector details: Appendix~\ref{app:trueloop}.}
\label{fig:loops-hurt}
\vspace{-5mm}
\end{figure}

Within each model's compressed setting, looping runs are also more likely
to fail than loop-free runs (Figure~\ref{fig:loops-hurt}).
The gap appears in all six Qwen3 and Gemini models: Qwen3-8B fails
\bad{85\%} of looping runs against 68\% of loop-free runs.
Gemini 3.1 Pro shows a wider gap: \bad{48\%} against 12\%.
A loop thus marks a run that is more likely to fail and has a higher
execution cost.

\subsection{Compression Makes Loops More Common}
\label{sec:compression}\label{sec:universality}
\begin{figure}[t]
\centering
\includegraphics[width=\linewidth]{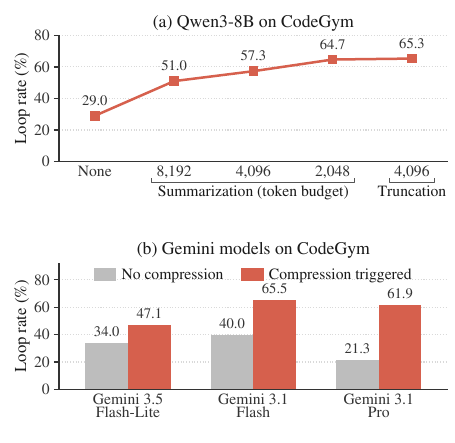}
\vspace{-8mm}
\caption{\textbf{Loop rates with and without compression.}
(a)~Qwen3-8B on CodeGym across context settings.
(b)~Gemini full-history runs versus compression-triggered episodes of summarization runs.}
\label{fig:compression}
\vspace{-5mm}
\end{figure}

An agent's interaction history records what it has already done.
To keep long rollouts within a context budget, the agent summarizes earlier
interactions (Appendix~\ref{app:setup}). We examine whether this compression
makes the agent more likely to repeat completed actions.
On CodeGym, Qwen3-8B loops in 29.0\% of runs with the full history.
Summarization at 4,096 tokens raises the loop rate to \bad{57.3\%}.
Lowering the threshold to 2,048 tokens raises it further to \bad{64.7\%}
(Figure~\ref{fig:compression}a).

The same pattern appears in all three Gemini models that we run with and without compression
(Figure~\ref{fig:compression}b). We compare CodeGym runs with the full
history against the subset of summarization runs in which compression fires.
Gemini 3.1 Pro loops least with the full history yet shows the largest
increase, \bad{2.9$\times$}.

To investigate this increase, we ask whether removing a completed action's
record makes the agent more likely to repeat it.
We reconstruct 300 histories from 253 CodeGym tasks after compression.
Each ends just before Qwen3-8B chooses its next action.
One version of each history keeps the most recent completed tool call and
its observation. The other removes this record.
We sample the next action eight times from each version.
The agent repeats that call in 3.3\% of samples with the record present
and \bad{48.8\%} with it removed
(Appendix~\ref{app:population-action-record}).

% The record therefore helps the agent avoid an immediate repeat.
% We next ask whether better context management can bring the loop rate back
% to the full-history level. We compare Qwen3-8B runs on CodeGym, measuring
% the fraction that contain at least one loop.
% Summaries designed to preserve past actions still
% leave substantial looping (Appendix Table~\ref{tab:battery}).
% Standard summarization yields a loop rate of 57.3\%.
% Alternative compressors ACON~\citep{acon} and LLMLingua-2~\citep{llmlingua2}
% reduce it to 46.3\%
% and 46.7\%, respectively. Both improve over standard summarization, but
% remain above the 29.0\% loop rate with the full history (Table~\ref{tab:main}).
% These results motivate training the policy itself to further reduce looping
% (Section~\ref{sec:method}).
The record therefore helps the agent avoid an immediate repeat. This explains
why compression causes more loops. However, over a whole episode, keeping the
record is not enough: Qwen3-8B still loops in 29.0\% of runs with the full
history. Under compression, a quarter of runs (25.2\%) repeat a call
that is still in the context word for word (Appendix~\ref{app:checks:context}).

Better context management does not close this gap. Summaries that keep an
explicit record of past actions, such as a trace of every executed call or a
ledger of tried actions, still loop in 52.0\% to 62.7\% of runs, close to
standard summarization at 57.3\% (Appendix Table~\ref{tab:battery}). The
compressors ACON~\citep{acon} and LLMLingua-2~\citep{llmlingua2} lower the
loop rate to 46.3\% and 46.7\%, still well above the 29.0\% with the full
history. The agent often fails to use a record it has, so we train the policy
itself (Section~\ref{sec:method}). This complements better compression:
DeLoop-Mix lowers the loop rate under every memory policy we test
(Section~\ref{sec:robustness}).
% blocks_analysis.tex -- Section 2 (Action-State Collapse) main-text floats, SPINE_v2 §3.
% Figures are built at final size by figures/v2/make_analysis_figs.py (values read from macros_gold.tex);
% include at \linewidth / \textwidth, never rescale. Place \input{figures/v2/blocks_analysis} near the
% start of sections/plain_02_analysis.tex.
%   fig:loops-hurt   single column (3.03 in)  -> Sec. 2.2
%   fig:compression  single column (3.03 in)  -> Sec. 2.3; panels (a) severity sweep, (b) Gemini
%                    full-width alternative: figures/v2/fig_compression_wide (6.3 x 1.85 in); to use it,
%                    switch to figure* and \includegraphics[width=\textwidth]{figures/v2/fig_compression_wide}.
% Model names: AUTHOR DECISION 2026-10-08 -- the 51%/24% Gemini cell is Gemini 3.1 Flash (as on slide 8; the source scripts
% build_slidefig_pctfail.py / build_slidefig_v5_failure_assoc.py and slide 7 say "Gemini 3 Flash (Dec 2025)"),
% so Figures 2 and 3 use one name (author check, notes/GOLD_NUMBERS.md #12). No benchmark is named in the
% fig:loops-hurt caption: the slide-8 sources are per-model summarize-arm autopsies (GOLD_NUMBERS / review N9).

% plain_04_method.tex — SPINE_v2 §2 row 5 (2026-10-08), rewritten from sections/plain_03_method.tex.
% The loop definition and detector now live in Section~\ref{sec:definition}; this section only uses them.
% Labels kept for old appendix references: sec:method, sec:diagnosis (mining subsection), sec:cure-method (training subsection).
% Every number enters via a \g macro (macros_gold.tex). Pair example: the Section 2.1 episode, verified in runtime/smoke/nf8k_smoke32.jsonl
% (8k corpus sample, task_index 252, rejected FindUnvisitedOnlineServer(3,0) -> chosen MarkVisited(3,0)); the slide-16
% Done(answer=1) example moved to the appendix (plain_appendix_notes.tex).
\section{Repeated Actions as Self-Supervision}
\label{sec:method}
% blocks_method.tex (2026-10-10): Section 4 method figure (fig:mining), built by figures/v2/make_fig_mining.py.
% The chosen-action rule is described neutrally ("the agent's own later loop-breaking action") until the
% author decides whether to state the mining code's rule (first later step whose observation is new).
\begin{figure}[t]
\centering
\includegraphics[width=\linewidth]{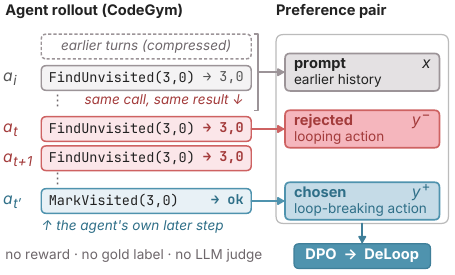}
\caption{\textbf{A self-mined preference pair.} Abridged Qwen3-8B CodeGym rollout
(tool names shortened; Appendix~\ref{app:real-loop}). $x$: history before the looping action;
$y^{-}$: looping action; $y^{+}$: first later loop-breaking action.}
\label{fig:mining}
\vspace{-4mm}
\end{figure}

To reduce looping under compressed context, we propose \textbf{DeLoop},
preference optimization on self-mined loop pairs. A detected looping action
serves as a negative example. If the same rollout later contains a different,
non-looping action, the first such action serves as a positive example
(Figure~\ref{fig:mining}).

\subsection{Mining Preference Pairs from Loops}
\label{sec:diagnosis}
We run the compressed agent on training tasks and apply the structural
detector of Section~\ref{sec:definition} to each rollout. For each detected
looping action $a_t$, we use that action as the rejected response $y^{-}$.
If the agent later recovers from the loop in the same rollout, we use the
first action that breaks the loop as the chosen response $y^{+}$.

For training, we pair both actions with the history $x=h_{<t}$ that preceded
the looping action:
\begin{gather*}
x=h_{<t},\qquad y^{-}=a_t,\qquad y^{+}=a_{t'},\\
t'=\min\{\,u>t : s_u=0,\ \kappa(a_u)\neq\kappa(a_t)\,\}.
\end{gather*}
If the agent does not recover from the loop before the rollout ends, we do
not construct a training pair for that looping action. A single rollout can
contribute multiple pairs.

For instance, the episode in Appendix~\ref{app:real-loop} yields one of our
self-mined pairs. The rejected action is the looping
\texttt{FindUnvisitedOnlineServer(\allowbreak{}start\_row=3, start\_col=0)}.
The chosen action is the agent's own later \texttt{MarkVisited(row=3, col=0)},
which breaks the loop.

Such a recovery provides a training pair even if the rollout ultimately
fails the task. We identify the looping action and the subsequent recovery
from the actions and observations alone. Pair construction therefore
requires no task reward, gold label, or LLM judge.

\subsection{Preference Training}
\label{sec:cure-method}
We train the policy $\pi_\theta$ to prefer the chosen action over the looping
action under the same context $x$. We use direct preference optimization
(DPO;~\citealp{dpo}), with the frozen base policy $\pi_{\mathrm{ref}}$ as the
reference:
\begin{equation*}
\mathcal{L}=-\log\sigma\!\Big(\beta\log\tfrac{\pi_\theta(y^{+}\mid x)}{\pi_{\mathrm{ref}}(y^{+}\mid x)}
-\beta\log\tfrac{\pi_\theta(y^{-}\mid x)}{\pi_{\mathrm{ref}}(y^{-}\mid x)}\Big).
\end{equation*}
For Qwen3-8B, we train for one epoch with LoRA DPO ($\beta=0.05$) on
8,000 self-mined CodeGym pairs. For Qwen3-14B and Qwen3-32B, we use the same
mining rule and $\beta$, with pairs mined from each model's own rollouts
(\cref{app:cure:recipe,app:cure:crossmodel}).

At inference time, we use the trained policy throughout the rollout to
generate each next action. In our standard setting, the harness triggers
compression when the estimated history exceeds 4,096 tokens. The same
trained model summarizes older interactions under a separate prompt. It
then continues acting from the updated context (Appendix~\ref{app:setup}).
DPO directly trains action preferences. Summary generation shares the
updated weights but has no separate training objective.

% plain_05_exp_a.tex: Section 5.1 setup, 5.2 main results, and 5.3 ablations.
% v2 final edit (notes/REVIEW_LOG_v2.md): scale claims rephrased (gains +7.2 at 14B, +7.1 at 32B are not monotone),
% DPO scoped to the four shown objectives (SimPO is appendix-only), GRPO described as online, "estimated" kept
% only in the Setup metric definition, recipe restatement and filler removed.
% Opens \section{Experiments}; plain_05_exp_b.tex continues with 5.4,
% and plain_06_beyond.tex continues with 5.5-5.6 within the same section.
% Result floats are input beside their corresponding subsections; fig:objectives stays in the appendix.
% 2026-10-08: results rewritten around the expanded Table 1, the
% restored fig:scale and the new tab:ablation (notes_main_additions_plan_20261008.md, sections A-C).
% Every result number is a \g macro from macros_gold.tex; 4,096 tokens and the
% 300 held-out tasks are settings, not results. Positive voice only: run counts,
% the 14B token-checkpoint provenance, early stopping, chars/4 token estimates,
% live in Limitations / the appendix (KTO/IPO/GRPO success values are now shown in tab:ablation).
\section{Experiments}
\label{sec:experiments}
We first evaluate DeLoop on CodeGym and examine its training signal.
We then vary the training data, context settings, and agent environments.
% Queue the wide main-result floats early so they stay near Section 5.2.
% blocks_results.tex -- main results: tab:main and fig:scale (Sec. 5.2).
% tab:ablation is in blocks_ablation.tex; fig:robust is in blocks_robustness.tex.
% Figures are built at final size (6.3 in wide) by
% figures/v2/make_result_figs.py from macros_gold.tex; include at \textwidth, never rescale.
% Tables: tables/v2/tab_main.tex, tables/v2/tab_ablation.tex (tabular only; captions live here).
% v2 final edit: tab:envs uses the single-column coltable environment defined in main.tex.
% Needs: booktabs, array, xcolor[table] (colortbl), TableHead/TableBase/TableOurs colors (all in main.tex).

\begin{table*}[t]
\centering
\footnotesize
\setlength{\tabcolsep}{3pt}
% tab:main (Sec. 5.1). Qwen3-8B, CodeGym-300, summarization at 4,096 tokens. Hand-written; every number is a
% macro: \g macros from macros_gold.tex (provenance in their % comments), plus \stateAct*, \guardDupBlock*,
% \guardRepPenalty*, \feedDistillEight* from macros.tex (L1658-1661, L1732-1746, L1298-1309).
% 2026-10-08 expansion (notes_main_additions_plan_20261008.md, section A): one table* with blocks for the
% full-history reference, training-free context methods, training-free inference-time controls, trained baselines with
% and without labels, and ours. Tokens per task are shown in thousands (one decimal; "TokK" macros).
% Bold = best value in a column, only where that best value is one of our rows: loop and steps (Ours),
% tokens (Mixed-data policy (ours), i.e. ours + cross-domain pairs). Success is not bolded (best is the StateAct-style tracker).
% Single-run rows carry no marker here; run counts live in sections/plain_appendix_notes.tex.
% 2026-10-08 review pass: Supervision cells of the reference/training-free rows read "none" (block headers say
% training-free); mixed-data row tokens/steps come from the efficiency table (macros.tex L882, L885).
% Colors: TableHead/TableGroup come from main_naacl.tex; the \providecolor lines keep the file compilable alone.
\providecolor{TableHead}{HTML}{EEF2F5}\providecolor{TableGroup}{HTML}{F6F7F8}%
\definecolor{TableBaseRose}{HTML}{FBE3E6}\definecolor{TableOursTeal}{HTML}{D7E7EF}%
\resizebox{\textwidth}{!}{%
\begin{tabular}{@{}llcccc@{}}
\toprule
\rowcolor{TableHead}
Method & Supervision (labels needed) & Loop (\%) $\downarrow$ & Success (\%) $\uparrow$ & Steps $\downarrow$ & Tokens/task (k) $\downarrow$ \\ \midrule
\rowcolor{TableGroup}\multicolumn{6}{l}{\textit{Reference}} \\
Full history & none & 29.0 & 31.7 & 12.23 & 24.3 \\
\rowcolor{TableGroup}\multicolumn{6}{l}{\textit{Training-free context methods}} \\
\rowcolor{TableBaseRose}Compressed baseline & none & 57.3 & 25.1 & 15.00 & 23.9 \\
Retrieval memory & none & 42.7 & 27.7 & 13.34 & 26.1 \\
ACON compressor & none & 46.3 & 29.3 & 14.84 & 20.5 \\
LLMLingua-2 compressor & none & 46.7 & 23.7 & 14.48 & 26.8 \\
History truncation & none & 65.3 & 18.0 & 16.05 & 34.8 \\
\rowcolor{TableGroup}\multicolumn{6}{l}{\textit{Training-free inference-time controls}} \\
Exact duplicate blocker & none & 46.7 & 25.3 & 14.71 & 24.9 \\
Reflexion on stuck & none & 55.3 & 24.3 & 14.62 & 24.1 \\
\rowcolor{TableGroup}\multicolumn{6}{l}{\textit{Trained with labels}} \\
Outcome DPO & outcome-ranked pairs (task success) & 34.7 & 25.7 & 10.99 & 15.9 \\
SFT on solved traces & solved trajectories (task success) & 42.7 & 21.3 & 10.61 & 25.2 \\
Gold-label DPO & looping pairs from outcomes (task success) & 22.3 & 25.7 & 8.10 & 11.3 \\
Gold-action GRPO & gold-action match reward (gold action) & 52.7 & 25.0 & 14.23 & 21.0 \\
\rowcolor{TableGroup}\multicolumn{6}{l}{\textit{Trained without labels}} \\
Distillation SFT & full-history traces (none) & 43.7 & 26.3 & 11.52 & 26.6 \\
\rowcolor{TableGroup}\multicolumn{6}{l}{\textit{Ours}} \\
\rowcolor{TableOursTeal}\textbf{DeLoop (ours)} & \textbf{self-mined pairs (none)} & \good{6.1}\impr{$-$51.2} & 25.9\impr{+0.8} & \good{4.67}\impr{$-$10.33} & 12.7\impr{$-$11.2} \\
\rowcolor{TableOursTeal}\textbf{DeLoop-Mix (ours)} & + cross-domain pairs (none) & 10.1\impr{$-$47.2} & \good{30.3}\impr{+5.2} & 5.94\impr{$-$9.06} & \good{9.6}\impr{$-$14.3} \\
\bottomrule
\end{tabular}}

\caption{\textbf{Main comparison on CodeGym.} Qwen3-8B under summarization at 4,096 tokens unless noted.
Steps and estimated tokens are per-task averages. Shading: our methods;
green: best compressed result; parentheses: change from the compressed baseline.}
\label{tab:main}
\vspace{-3mm}
\end{table*}

\begin{figure*}[t]
\centering
\includegraphics[width=\textwidth]{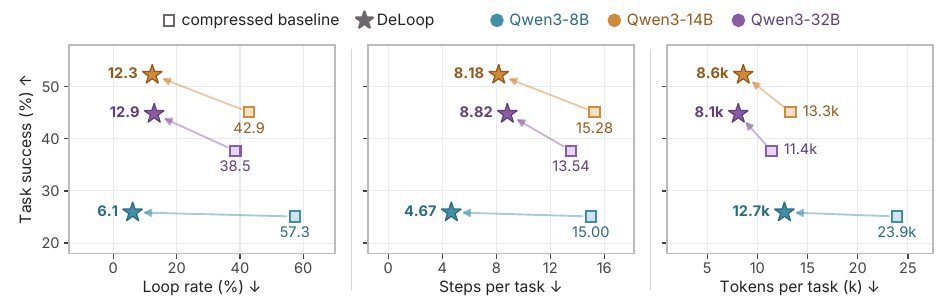}
\caption{\textbf{Model scaling on CodeGym.} Qwen3-8B, Qwen3-14B, and Qwen3-32B under
summarization at 4,096 tokens. Task success versus loop rate, steps, and estimated tokens per task
(left to right). Squares: compressed baseline; stars: DeLoop; colors: model size.}
\label{fig:scale}
\vspace{-6mm}
\end{figure*}

\subsection{Experimental Setup}
\label{sec:setup-main}
\paragraph{Benchmarks and models.}
CodeGym~\citep{du2025generalizable} is our primary benchmark for both mining
and evaluation: its tasks require many act--observe rounds, so the history
grows until compression fires. We also evaluate on four further agent
environments: ALFWorld~\citep{alfworld}, Open-AgentRL~\citep{openagentrl},
AppWorld~\mbox{\citep{appworld}}, and ScienceWorld~\citep{scienceworld}.
Terminal-Bench~\citep{terminalbench} and BrowseComp-Plus~\citep{browsecompplus}
provide further evaluations in their native agent harnesses
(Section~\ref{sec:transfer-bench}). Our main agent is Qwen3-8B, and we
repeat the CodeGym study with Qwen3-14B and Qwen3-32B~\citep{qwen3}.

\paragraph{Training variants.}
DeLoop trains on self-mined CodeGym pairs (Section~\ref{sec:cure-method}).
We also evaluate \textbf{DeLoop-Mix}, a variant trained with additional
cross-domain pairs (Section~\ref{sec:cross-task}). Its training data include
pairs mined from Terminal-Bench rollouts.

\paragraph{Evaluation protocol and metrics.}
In our standard
compressed setting, the agent summarizes its history whenever it exceeds
4,096 tokens; the same model writes the summary and continues from it. Unless
noted otherwise, each policy in the main CodeGym comparison acts in this
setting on 300 held-out tasks, and the full-history agent serves as the
uncompressed reference. We
report four metrics: the loop rate (Section~\ref{sec:analysis}), task
success, steps per task, and estimated tokens per task
(Appendix~\ref{app:notes}).

\paragraph{Baselines.}
Training-free context methods change what the agent
keeps from its history, including retrieval memory, ACON~\citep{acon}, and
LLMLingua-2~\citep{llmlingua2}. Training-free inference-time controls act during
inference, including an exact duplicate blocker and
Reflexion~\citep{reflexion}. Trained baselines learn from task outcomes, solved
trajectories, or gold actions, and distillation SFT on full-history traces
is, like DeLoop, label-free. We do not run GRPO on task success: its reward
arrives only once per episode, and even a per-step gold-action reward leaves
only 2.7\% of GRPO groups informative (Appendix~\ref{app:cure:grpo-diagnosis}).
The full battery is in Appendix
Tables~\ref{tab:battery} and~\ref{tab:inference-controls}.
% \tianle{We can add more details about the labeled training if needed} [commented out 2026-10-10]

\subsection{Main Results on CodeGym}
\label{sec:cure}\label{sec:scale}
We first test whether DeLoop reduces loops and interaction cost while
maintaining task success on CodeGym (Table~\ref{tab:main}).

\paragraph{Qwen3-8B results.}
Self-mined pairs give the lowest loop rate of all methods. DeLoop
cuts the loop rate \good{9.4$\times$}, from 57.3\% to
\good{6.1\%}. Training-free methods in Table~\ref{tab:main} remain between 42.7\% and 65.3\%,
and the strongest labeled baseline, gold-label DPO, reaches 22.3\%.
Breaking the loop also removes wasted interaction: steps per task fall from
15.00 to \good{4.67}, fewer than any other method, and tokens
per task fall by \good{47\%}, while task success holds.

\paragraph{Model scaling.}
We next test the same training signal at larger model sizes.
We mine pairs from the compressed CodeGym rollouts of Qwen3-14B and
Qwen3-32B with the same rule and train each model with DPO
(Figure~\ref{fig:scale}). Loops fall at every scale,
\good{3.5$\times$} on Qwen3-14B and
\good{3.0$\times$} on Qwen3-32B. Tokens per task fall by \good{35\%} and
\good{29\%}, respectively. At these scales the repair also raises task success, by
\good{7.2} points on Qwen3-14B and \good{7.1} points
on Qwen3-32B.

\paragraph{General capability.}
We also check capabilities beyond agent rollouts. Performance is preserved
on LiveCodeBench, MT-Bench, and IFEval
(Appendix~\ref{app:cure:livecodebench}).

\subsection{Ablation Studies}
\label{sec:objectives}
% tab:ablation, placed with Section 5.3. Table data are unchanged.
\begin{coltable}[!t]
\centering
\small
\setlength{\tabcolsep}{3pt}
% tab:ablation (Sec. 5.3), single-column (coltable in figures/v2/blocks_results.tex). Added 2026-10-08 per
% notes_main_additions_plan_20261008.md, section C. Qwen3-8B, CodeGym-300, summarization at 4,096 tokens; every
% trained row uses the same 8k self-mined pair pool (GRPO uses the same looping-action flag as a dense penalty).
% Values: \gBaseSumm*, \gObj* (macros_gold.tex L103-105, L186-200) and \gSig* (macros_gold.tex, Ablation block;
% source tables/tab_signal_decomposition_appendix.tex L10-11). 2026-10-08 review pass: Steps column removed (the
% pair-signal source counts rounds, not steps; note in plain_appendix_notes.tex). Bold = best per column (ours).
\providecolor{TableHead}{HTML}{EEF2F5}\providecolor{TableGroup}{HTML}{F6F7F8}%
\providecolor{TableBaseRose}{HTML}{FBE3E6}\providecolor{TableOursTeal}{HTML}{D7E7EF}%
\begin{tabular}{@{}lcc@{}}
\toprule
\rowcolor{TableHead}
Policy & Loop (\%) $\downarrow$ & Success (\%) $\uparrow$ \\ \midrule
\rowcolor{TableBaseRose}Compressed baseline & 57.3 & 25.1 \\
\rowcolor{TableGroup}\multicolumn{3}{l}{\textit{Objective (same loop signal)}} \\
\rowcolor{TableOursTeal}\textbf{DPO (DeLoop)} & \good{6.1}\impr{$-$51.2} & \good{25.9}\impr{+0.8} \\
KTO & 12.6 & 23.8 \\
IPO & 11.7 & 24.0 \\
Looping-action penalty GRPO & 15.0 & 24.7 \\
SFT (chosen action only) & 37.7 & 21.3 \\
\bottomrule
\end{tabular}

\caption{\textbf{Training-objective ablations.} Qwen3-8B on CodeGym under summarization at 4,096 tokens.
DPO, KTO, IPO, and SFT use the same 8,000 self-mined pairs; SFT uses only chosen actions.
GRPO penalizes looping actions in online rollouts. Parentheses: change from the compressed baseline.}
\label{tab:ablation}
\vspace{-4mm}
\end{coltable}

We next examine the roles of the preference pair and the training objective
(Table~\ref{tab:ablation}).

\paragraph{Contribution of each side of the pair.}
The looping action matters. Imitating only the loop-breaking action
leaves a loop rate of 37.7\%, while the contrast with the
looping action brings it to \good{6.1\%}.

\paragraph{Alternative training objectives.}
We then test the loop signal beyond DPO, using KTO~\citep{ethayarajh2024kto},
IPO~\citep{azar2024ipo}, and online GRPO~\citep{shao2024deepseekmath}. Every objective cuts the loop rate to \good{15.0\%} or below, and
DPO (DeLoop) gives the lowest loop rate (\good{6.1\%}) and the highest task success.

\paragraph{Remaining repeated actions.}
We also examine which repeated actions remain after training.
The share of repeats that return new information rises from
29.9\% to \good{69.8\%} (Appendix~\ref{app:cure:selectivity}).

% plain_05_exp_b.tex: Section 5.4, the effect of cross-domain training data.
% v2 final edit (notes/REVIEW_LOG_v2.md): no implied "largest gains" comparatives (other success cells are negative,
% see Limitations); count macros spelled out as words; per-environment checkpoints described as DPO on self-mined loop
% pairs ... -- HOLD (author decision pending, 2026-10-08): main text uses the slide wording 'our DPO checkpoint per environment';
% the per-environment checkpoint attribution is commented out in Limitations and plain_appendix_notes.tex until the authors decide.
% Continues \section{Experiments}; the shared robustness figure is input here
% for Sections 5.4-5.5. Every number is a \g macro
% from macros_gold.tex; context budgets (2,048 / 4,096 / 8,192 tokens) are
% settings, not results. Positive voice only: negative success cells, run
% counts, and the per-environment recipes
% live in Limitations / the appendix.

\subsection{Effect of Cross-Domain Training Data}
\label{sec:cross-task}
% fig:robust, shared by Sections 5.4 (panel a) and 5.5 (panels b-c).
\begin{figure*}[t]
\centering
\includegraphics[width=\textwidth]{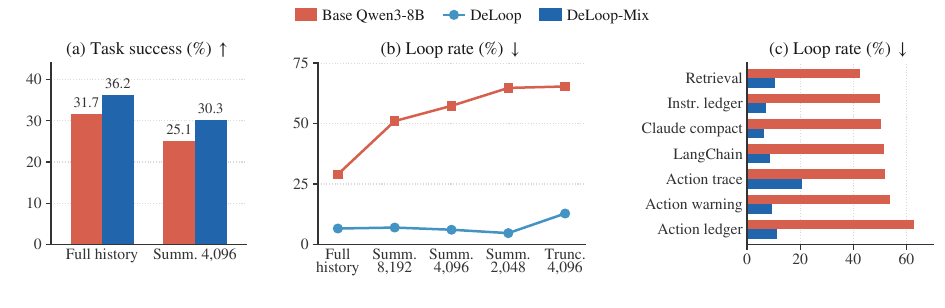}
\vspace{-10mm}
\caption{\textbf{Cross-domain data and context management.} Qwen3-8B on CodeGym.
\textit{(a)} Task success with and without compression.
\textit{(b)} Loop rate across context settings.
\textit{(c)} Loop rate across memory policies. Base: untrained policy.}
\label{fig:robust}
\vspace{0mm}
\end{figure*}

Self-mined CodeGym pairs preserve task success at 8B. We next test whether
adding experience from other agent tasks can improve it.
We keep the 8,000 CodeGym pairs fixed, add
4,635 cross-domain pairs from ALFWorld,
Open-AgentRL, and Terminal-Bench
rollouts, mined with the same label-free rule, and train Qwen3-8B on all
12,635 pairs with the same DPO recipe to obtain \textbf{DeLoop-Mix}.

DeLoop-Mix raises success both with and without
compression (Figure~\ref{fig:robust}a). Under summarization at 4,096 tokens,
the CodeGym success of Qwen3-8B rises from 25.1\% to
\good{30.3\%}, a gain of \good{5.2} points, compared with
+0.8 points for DeLoop. With the full
history, success rises from 31.7\% to \good{36.2\%}, a gain of
4.5 points. The loop repair stays in place: the loop rate
falls from 29.0\% to 10.1\% without compression and from
57.3\% to \good{10.1\%} with it.
Under summarization, DeLoop-Mix also uses the fewest tokens per task shown in
Table~\ref{tab:main}.

% plain_06_beyond.tex (2026-10-08): new Section 6 at the authors' request -- the old Sec. 5.5 (compression
% settings and memory policies) and Sec. 5.6 (other agent environments), text unchanged. Table 2 (tab:envs)
% is unchanged and now floats with this section (figures/v2/blocks_beyond.tex).
\section{Robustness and Generalization}
\label{sec:beyond}
The CodeGym results show that self-mined pairs reduce loops and interaction
cost. We now test whether these gains extend to other context settings,
memory policies, and agent environments.
% tab:envs float, moved from blocks_results.tex with the new Section 6 (content unchanged).
\begin{coltable}[t]
\centering
\footnotesize
\setlength{\tabcolsep}{2.3pt}
\resizebox{\linewidth}{!}{% tab:envs (SPINE_v2 §3, Sec. 5.6). Qwen3-8B, base with summarization vs our DPO checkpoint per environment,
% native loop and success metrics. Hand-written; every number is a \g macro from macros_gold.tex
% (provenance: notes/GOLD_NUMBERS.md "Result 5"; Open-AgentRL tokens are not recorded).
% v2 final edit: no bold (Ours columns are shaded); single-column float (coltable in blocks_results.tex).
% Ours columns tinted with TableOurs (main.tex; \providecolor keeps the file standalone).
% 2026-10-11: Ours = DeLoop-Mix (three-source mix DPO) in every row, per author. Open-AgentRL: 3-run base (.7898/.4391) vs one DeLoop-Mix run (.7351/.4552), training log 2026-08-10.
% 2026-10-13, per author: ALFWorld row = mix3nl (big_mix3nlalf_summarize; same setting as mix3dpo, no filtering): loop .276 /
% success .381 (macros alfMixThreeNoLabel*), tokens 6.2k (tools/efficiency_metrics.py; the mix3dpo row recomputed to 7.6k as a check).
% Previous ALFWorld row (mix3dpo, big_mix3dpoalf_summarize): 47.8 / 27.6 / 7.6k.
% ALFWorld tokens (old) 7.6k = 4.4k input + 3.2k output (tab_mix_zero_label_appendix); ScienceWorld 6.0k recomputed with tools/efficiency_metrics.py (mix3nl check: 5.2k).
% v3: Ours columns in the light steel-teal tint of the figure palette (notes/STYLE_v3_AUQ.md).
\definecolor{TableOursTeal}{HTML}{D7E7EF}%
\begin{tabular}{@{}lc>{\columncolor{TableOursTeal}}cc>{\columncolor{TableOursTeal}}cc>{\columncolor{TableOursTeal}}c@{}}
\toprule
 & \multicolumn{2}{c}{Loop (\%) $\downarrow$} & \multicolumn{2}{c}{Success (\%) $\uparrow$} & \multicolumn{2}{c}{Tokens (k) $\downarrow$} \\
\cmidrule(lr){2-3}\cmidrule(lr){4-5}\cmidrule(lr){6-7}
Env. & Base & Ours & Base & Ours & Base & Ours \\ \midrule
ALFWorld & 55.2 & \good{27.6}\impr{$-$27.6} & 37.3 & \good{38.1}\impr{+0.8} & 9.3 & \good{6.2}\impr{$-$3.1} \\
Open-AgentRL & 79.0 & \good{73.5}\impr{$-$5.5} & 43.9 & \good{45.5}\impr{+1.6} & -- & -- \\
AppWorld & 73.7 & \good{40.9}\impr{$-$32.8} & 7.0 & 5.8\decl{$-$1.2} & 17.9 & \good{14.1}\impr{$-$3.8} \\
ScienceWorld & 96.5 & \good{83.5}\impr{$-$13.0} & 3.5 & 1.5\decl{$-$2.0} & 6.5 & \good{6.0}\impr{$-$0.5} \\
\bottomrule
\end{tabular}
}
\caption{\textbf{Evaluation in four additional agent environments.} Base: untrained Qwen3-8B. Ours:
DeLoop-Mix. Native metrics; green: improvement; parentheses: change from Base.}
\label{tab:envs}
\vspace{-4mm}
\end{coltable}

\subsection{Generalization Across Compression Settings and Memory Policies}
\label{sec:robustness}
We first evaluate
our Qwen3-8B DeLoop policy on CodeGym under five context settings, from
the full history to summarization at 2,048 tokens and truncation at 4,096
tokens. The loop rate falls in every setting, remaining between 4.7\% and
12.8\% while the untrained baseline reaches 65.3\% (Figure~\ref{fig:robust}b).

Beyond the amount of retained history, we vary how it is represented.
We run DeLoop-Mix (\cref{sec:cross-task}) under seven
alternative memory policies, from retrieval memory and Claude-style
compaction to summaries augmented with an action ledger. The loop rate,
steps per task, and tokens per task fall under all seven
(Figure~\ref{fig:robust}c); with the action ledger,
the loop rate falls from 62.7\% to
\good{11.3\%} and task success rises by
\good{5.7} points.

\subsection{The Improvement Extends to Other Agent Environments}
\label{sec:generalization}
The context comparisons above all use CodeGym. We next run DeLoop-Mix in
four further environments,
ALFWorld, Open-AgentRL,
AppWorld, and ScienceWorld, scored with
each environment's native loop and success metrics (Table~\ref{tab:envs}). ALFWorld and Open-AgentRL
are evaluated on unseen games, disjoint from the training games behind their pairs
(Appendix~\ref{app:cure:recipe}). Loops fall in all four, by up
to 32.8 points on AppWorld. Tokens per task fall wherever recorded, and task
success on Open-AgentRL rises from 43.9\% to \good{45.5\%}.

\subsection{The Repair Transfers to New Benchmarks}
\label{sec:transfer-bench}
Finally, we test DeLoop and DeLoop-Mix in native agent harnesses with different
tools and interactions. Terminal-Bench tests
terminal tasks, and BrowseComp-Plus tests web research.
On Terminal-Bench,
DeLoop cuts unproductive duplicate calls from 16.6 to \good{6.9}
per task. On all 830 BrowseComp-Plus queries,
DeLoop-Mix lowers the loop rate from 22.4\% to \good{1.2\%}
while matching the untrained model's accuracy;
DeLoop, trained on CodeGym pairs alone, already lowers it to
6.4\%. We also test zero-shot transfer from CodeGym to ALFWorld.
DeLoop on Qwen3-32B, trained only on CodeGym pairs, lowers the ALFWorld loop rate from
64.9\% to \good{54.5\%} and raises task success from
33.6\% to \good{42.5\%}.

% \tianle{Mentioning the training on other benchmarks} [commented out 2026-10-10]

% plain_06_conclusion.tex (2026-10-08, SPINE_v2 §2 row "6 Conclusion").
% Positive voice only; every number is a \g macro from macros_gold.tex. Caveats live in plain_07_limitations.tex.
\section{Conclusion}
\label{sec:conclusion}
\label{sec:discussion}

Long-horizon agents that compress their history fall into loops: they
re-issue an action they already ran and receive the same result. On
CodeGym, summarizing the history raises the loop rate of Qwen3-8B from
29.0\% to 57.3\%, and better context management alone does
not remove it. Our key observation is that the loop labels itself: the
looping action is the rejected action, and the agent's later loop-breaking
action is the chosen one. Trained on these self-mined pairs, DeLoop cuts the loop rate
9.4$\times$, from 57.3\% to 6.1\%, at about one third of the steps per task while
maintaining task success, and raises task success by 7.2 points at 14B and
7.1 points at 32B. An agent's own loops become its supervision: break the
loop, keep the progress. We believe our work could help the community better
understand and reduce loops in long-horizon agents.

% Google report: no Limitations section (it is required only for the ACL submission; see main_naacl.tex).

\bibliography{refs}
\clearpage
\appendix
% Appendix layout (as in the NAACL version): no stretched columns, fuller float pages, no forced float flush.
\raggedbottom
\renewcommand{\topfraction}{0.9}\renewcommand{\dbltopfraction}{0.9}\renewcommand{\bottomfraction}{0.8}\renewcommand{\textfraction}{0.05}
\renewcommand{\floatpagefraction}{0.75}\renewcommand{\dblfloatpagefraction}{0.75}
\setcounter{topnumber}{4}\setcounter{dbltopnumber}{4}\setcounter{bottomnumber}{2}\setcounter{totalnumber}{8}
\let\FloatBarrier\relax
% Appendix table of contents: lists only the appendix sections and subsections.
\iflatexml\else
\section*{Appendix Contents}
\startcontents[appendix]
\printcontents[appendix]{}{1}{\setcounter{tocdepth}{2}}
\fi
% plain_appendix_notes.tex (rewritten 2026-10-08): the evaluation details the main text points to
% (run counts, token estimates, loop measurement). Context settings live in app:setup (review 2026-10-08). Caveats formerly parked here
% now live in Limitations (sections/plain_07_limitations.tex).
% Label defined: app:notes.
\section{Evaluation Details}
\label{app:notes}

\begin{figure*}[t]
\centering
\includegraphics[width=\textwidth]{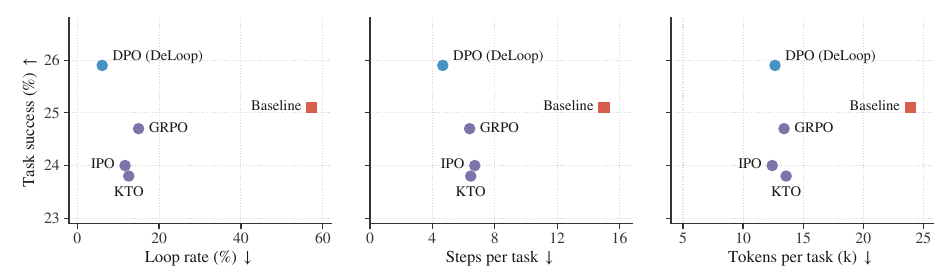}
\caption{\textbf{Training objectives on the loop signal.} Qwen3-8B on CodeGym under
summarization at 4,096 tokens. Task success versus loop rate, steps, and estimated tokens per task
(left to right). Red square: compressed baseline; blue: DeLoop; purple: KTO, IPO, and looping-action penalty GRPO.}
\label{fig:objectives}
\end{figure*}

\paragraph{Evaluation runs.}
Unless noted otherwise, every number comes from the 300 held-out CodeGym
tasks under summarization at 4,096 tokens. When a number averages several
evaluation runs, each run evaluates the same trained policy on the same tasks.
In
Table~\ref{tab:main}, the full-history agent, the compressed baseline,
DeLoop, and DeLoop-Mix each average three runs, history truncation
averages two, and every other row is one run. In Table~\ref{tab:ablation}
and Figure~\ref{fig:objectives}, DPO (DeLoop) and KTO average three runs, while IPO and
the looping-action penalty GRPO are one run each; the KTO tokens per task
come from one of its three runs. In Figure~\ref{fig:scale}, the Qwen3-14B
compressed baseline averages three runs and its DeLoop policy is one run; the
Qwen3-32B compressed baseline averages two runs and its DeLoop policy three. In
Figure~\ref{fig:robust}, the summarization cells at 8,192 and 2,048 tokens and
the seven memory policies are one run each. In Table~\ref{tab:envs}, each
cell is one run except ScienceWorld, the AppWorld DeLoop-Mix cells, and the
Open-AgentRL base, which average three.

\paragraph{Steps and tokens per task.}
Steps per task count the actions the agent executes in a task. We estimate
tokens per task by counting the logged trajectory with the Qwen tokenizer; the
trajectory holds the task prompt, the observations, and the model's outputs,
each counted once. 
% All token numbers in Sections~\ref{sec:experiments}
% and~\ref{sec:beyond}, in Table~\ref{tab:battery}, and in
% Figure~\ref{fig:loop-cost} use this estimate. 
Open-AgentRL does not record
tokens, so Table~\ref{tab:envs} leaves its token cells empty.

\begin{figure*}[t]
\centering
\includegraphics[width=\textwidth]{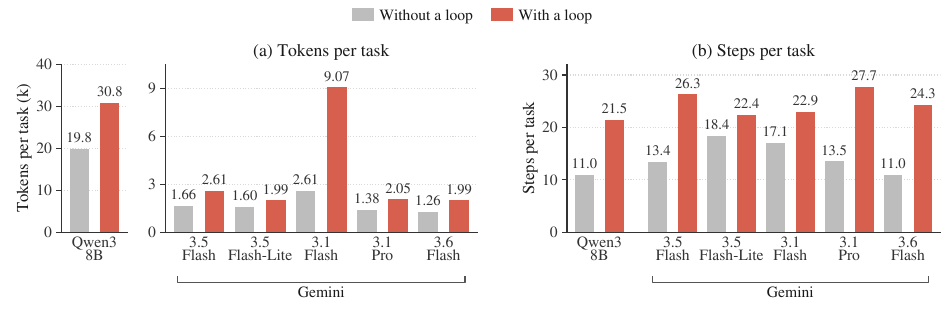}
\caption{\textbf{Interaction cost of looping and loop-free runs.}
(a)~Transcript tokens and (b)~steps per task for runs with a loop (red) and without one (gray),
for Qwen3-8B on CodeGym and five Gemini models.
Tokens use the Qwen tokenizer (Appendix~\ref{app:notes}); panel (a) uses a separate scale for Qwen3-8B.}
\label{fig:loop-cost}
\end{figure*}

\paragraph{Loop measurement.}
On CodeGym, the loop rate uses the structural detector of
Section~\ref{sec:definition} (Appendix~\ref{app:setup}); each environment in
Table~\ref{tab:envs} uses its own native repeat detector.
Figures~\ref{fig:loops-hurt} and~\ref{fig:loop-cost} compare runs within each
model's compressed setting; Appendix~\ref{app:trueloop} gives the runs
behind Figure~\ref{fig:loops-hurt}.

% plain_appendix_example.tex -- "A real loop" example, moved from Section 3.1 (plain_02_analysis.tex) on 2026-10-08;
% light rewrite 2026-10-08. The FindUnvisitedLand / Done(answer=1) pair moved here from plain_appendix_notes.tex.
% Label defined: app:real-loop. Labels referenced: sec:method.
\section{A Real Loop from CodeGym}
\label{app:real-loop}
The episode below is excerpted from a training rollout of the
compressed Qwen3-8B agent. The task is to count clusters of online servers on
a grid:
\begin{center}\scriptsize
\begin{tabular}{@{}l@{}}
\textit{$\cdots$ 13 earlier act--observe turns compressed $\cdots$}\\[1pt]
\texttt{FindUnvisitedOnlineServer(start\_row=3, start\_col=0)} $\rightarrow$ \texttt{3,0}\\[1pt]
\textit{$\cdots$ further act--observe turns of real work $\cdots$}\\[1pt]
\texttt{FindUnvisitedOnlineServer(start\_row=3, start\_col=0)} $\rightarrow$ \texttt{3,0}\\[1pt]
\texttt{FindUnvisitedOnlineServer(start\_row=3, start\_col=0)} $\rightarrow$ \texttt{3,0}\\[1pt]
\textit{later in the episode:}\quad \texttt{MarkVisited(row=3, col=0)}
\end{tabular}
\end{center}
The agent runs the same search three times and gets \texttt{3,0} back every
time: nothing has marked that server as visited, so the search can only
return the same cell. The loop sits inside an otherwise productive episode,
and the action the search was waiting for, \texttt{MarkVisited}, appears later
in the same rollout. This is the structure DeLoop mines
(Section~\ref{sec:method}): the repeated search is the looping action, and
\texttt{MarkVisited} is the loop-breaking action.

Loop-breaking actions range from tool calls that make progress to final
answers. Another self-mined pair rejects a looping
\texttt{FindUnvisitedLand()} call and chooses the agent's later
\texttt{Done(answer=1)}.

% blocks_analysis_appendix.tex -- appendix floats for Section 3 (captions rewritten 2026-10-08).
%   fig:loop-cost  full width (6.3 in), built by figures/v2/make_analysis_figs.py; slide 9 values
%                  (\gCost* macros are axis-calibrated reads of slide 9, see notes/GOLD_NUMBERS.md).
%   tab:battery    generated table tables/v2/tab_battery.tex (\gCtx* macros + the CodeGym-300 battery).
% main.tex redefines `table` as a two-column float, so tab:battery spans the page width.

\begin{table}[t]
\centering
\small
% AUTO-GENERATED by figures/v2/make_analysis_figs.py -- do not hand-edit.
% tab:battery. Qwen3-8B, CodeGym-300. Values: \gCtx* macros (macros_gold.tex) where they exist,
% otherwise tables/tab_compressor_battery_appendix.tex (source cited by notes/GOLD_NUMBERS.md).
% Every printed literal below was checked against that source at build time.
\definecolor{TableLoopRose}{HTML}{FCE9EB}% rose tint (notes/STYLE_v3_AUQ.md), local to this float
\setlength{\tabcolsep}{9pt}
\begin{tabular}{lccccc}
\toprule
\rowcolor{TableHead}
Context method & Loop rate (\%) $\downarrow$ & Success (\%) $\uparrow$ & Steps/task $\downarrow$ & Tokens/task $\downarrow$ & Fired (\%) \\ \midrule
No compression (full history) & 29.0 & 31.7 & 12.23 & 24,319 & 0.0 \\ \midrule
\rowcolor{TableGroup}\multicolumn{6}{l}{\textit{Summaries}} \\
\rowcolor{TableLoopRose}Standard rolling summary & 57.3 & 25.1 & 15.00 & 23,915 & 96.6 \\
Instruction-ledger summary & 50.0 & 27.0 & 12.40 & 19,535 & 95.7 \\
Claude-style compact summary & 50.3 & 26.3 & 13.67 & 20,944 & 95.0 \\
LangChain-style summary & 51.7 & 26.3 & 13.25 & 20,725 & 96.7 \\
Action-trace summary & 52.0 & 25.0 & 14.21 & 30,937 & 96.7 \\
Action-warning summary & 54.0 & 24.0 & 14.12 & 22,886 & 97.0 \\
Action-ledger summary & 62.7 & 23.3 & 15.51 & 24,509 & 96.7 \\
\rowcolor{TableGroup}\multicolumn{6}{l}{\textit{Memory and learned compressors}} \\
Retrieval memory & 42.7 & 27.7 & 13.34 & 26,127 & 97.0 \\
ACON compressor & 46.3 & 29.3 & 14.84 & 20,542 & 97.7 \\
LLMLingua-2 compressor & 46.7 & 23.7 & 14.48 & 26,788 & 97.0 \\
\rowcolor{TableGroup}\multicolumn{6}{l}{\textit{Truncation}} \\
Round clipping (keep 6 of 12) & 37.0 & 28.3 & 12.17 & 25,956 & 44.0 \\
History truncation & 65.3 & 18.0 & 16.05 & 34,794 & 96.7 \\
\bottomrule
\end{tabular}

\caption{\textbf{Context-management comparison.} Qwen3-8B on 300 held-out CodeGym tasks.
Fired: share of episodes in which compression fired. Steps and estimated tokens are per-task averages.
The shaded row is the compressed baseline: summarization at 4,096 tokens.}
\label{tab:battery}
\end{table}

% blocks_results_appendix.tex -- appendix result float (SPINE_v2 §3): fig:objectives (Result 6).
% Built at final size (6.3 in wide) by figures/v2/make_result_figs.py from macros_gold.tex; include at
% \textwidth, never rescale. SimPO is not shown (Limitations).

% fig:scale moved back to the main text (figures/v2/blocks_results.tex, Sec. 5.2) on 2026-10-08.

% plain_appendix_oldfloats.tex -- rewritten 2026-10-08 (appendix rewrite). Keeps the two old floats that add
% detail to the main text: compression methods on Gemini 3.5 Flash-Lite (tab:compressors-main, formerly in
% naacl_app/naacl_appendix_moved.tex) and per-model CodeGym results (tab:sizes, CodeGym rows of the former
% tab:generalization-matrix in naacl_app/appendix_transfer.tex). All other old floats were removed.
\section{More Compression Methods and Model Sizes}
\label{app:more-results}

\begin{coltable}[t]
\centering
\footnotesize
\setlength{\tabcolsep}{3pt}
% tab_compressors_main (naacl_app private copy), rewritten 2026-10-08: Gemini 3.5 Flash-Lite on CodeGym.
% Values copied from the earlier auto-generated table ("loop after compression" and "success"), shown as
% percentages. The summarization loop rate follows main-text Figure 3b (47.1; source 0.471).
% The no-compression loop rate is the all-episode rate, as in Figure 3b.
\begin{tabular}{@{}lcc@{}}
\toprule
\rowcolor{TableHead}
Context method & Loop (\%) $\downarrow$ & Success (\%) $\uparrow$ \\ \midrule
No compression (full history) & 34.0 & 57.0 \\ \midrule
Summarization & 47.1 & 54.7 \\
Truncation & 48.7 & 46.0 \\
Claude-style compact summary & 49.0 & 56.0 \\
LangChain-style summary & 49.7 & 49.7 \\
Structured action trace & 64.4 & 46.3 \\
\bottomrule
\end{tabular}

\caption{\textbf{Compression methods on Gemini 3.5 Flash-Lite.} Gemini 3.5 Flash-Lite on CodeGym. As in
Figure~\ref{fig:compression}b, the loop rate counts the episodes in which
compression fired (all episodes without compression); task success counts all
tasks.}
\label{tab:compressors-main}
\end{coltable}

\begin{coltable}[t]
\centering
\footnotesize
\setlength{\tabcolsep}{2pt}
\resizebox{\linewidth}{!}{% tab_generalization_matrix (naacl_app private copy), rewritten 2026-10-08 as the per-model CodeGym table
% (tab:sizes). Values copied from the CodeGym rows of the earlier auto-generated breadth matrix, shown as
% percentages and k tokens; the Qwen3-8B row is main-text Table 1. Columns styled like tables/v2/tab_envs.tex.
\definecolor{TableOursTeal}{HTML}{D7E7EF}%
\begin{tabular}{@{}lc>{\columncolor{TableOursTeal}}cc>{\columncolor{TableOursTeal}}cc>{\columncolor{TableOursTeal}}cc>{\columncolor{TableOursTeal}}c@{}}
\toprule
 & \multicolumn{2}{c}{Loop (\%) $\downarrow$} & \multicolumn{2}{c}{Success (\%) $\uparrow$} & \multicolumn{2}{c}{Steps $\downarrow$} & \multicolumn{2}{c}{Tokens (k) $\downarrow$} \\
\cmidrule(lr){2-3}\cmidrule(lr){4-5}\cmidrule(lr){6-7}\cmidrule(lr){8-9}
Model & Base & DeLoop & Base & DeLoop & Base & DeLoop & Base & DeLoop \\ \midrule
Qwen3-8B & 57.3 & \good{6.1} & 25.1 & \good{25.9} & 15.00 & \good{4.67} & 23.9 & \good{12.7} \\
Qwen3-14B & 42.9 & \good{12.3} & 45.1 & \good{52.3} & 15.28 & \good{8.18} & 13.3 & \good{8.6} \\
Qwen3-32B & 38.5 & \good{12.9} & 37.7 & \good{44.8} & 13.54 & \good{8.82} & 11.4 & \good{8.1} \\
Qwen3.5-9B & 54.7 & \good{26.0} & 49.7 & \good{55.3} & 19.15 & \good{14.95} & 8.0 & \good{5.8} \\
\bottomrule
\end{tabular}
}
\caption{\textbf{Per-model CodeGym results.}
CodeGym under summarization at 4,096 tokens: compressed baseline (Base) and
DeLoop trained on pairs mined from the same model's rollouts.
Tokens are estimated tokens per task. Shaded columns: DeLoop; green marks an
improvement over the baseline.}
\label{tab:sizes}
\end{coltable}

\paragraph{Every compression method raises looping on Gemini 3.5 Flash-Lite.}
Figure~\ref{fig:compression}b compares Gemini models with and without
summarization. Table~\ref{tab:compressors-main} repeats this comparison for
Gemini 3.5 Flash-Lite under five compression methods. Every method raises the
loop rate, from 34.0\% without compression to between 47.1\% and 64.4\%, and
every method lowers task success. The structured action trace loops most
often, so an explicit record of past actions does not stop the loop on
Gemini either.

\paragraph{The repair holds for every model we train.}
Table~\ref{tab:sizes} gives the CodeGym results behind
Figure~\ref{fig:scale}, together with steps per task and a fourth model,
Qwen3.5-9B. Steps per task fall at every size: from 15.00 to 4.67 on
Qwen3-8B, from 15.28 to 8.18 on Qwen3-14B, and from 13.54 to 8.82 on
Qwen3-32B. Qwen3.5-9B follows the same pattern: DeLoop cuts its loop
rate from 54.7\% to 26.0\% and raises task success by 5.6 points, from
49.7\% to 55.3\%.

% Setup, loop labels, and action-record probes (rewritten 2026-10-08 to match the main text).
% Caveats and negative results from the earlier version moved to the Limitations section.

\section{Setup Details}
\label{app:setup}

% Float for app:trueloop, placed early so the two-column float lands next to that section.
\begin{table}[t]
\centering
\small\setlength{\tabcolsep}{4pt}
% Loop labels and failure rates behind fig:loops-hurt (rewritten 2026-10-08).
% Source: drive_restore_20260821/agent_benchmarks/analysis/figs/build_slidefig_v5_failure_assoc.py
% (FAIL_GIVEN_LOOP / FAIL_GIVEN_NOLOOP) and the per-model 2x2 sources cited in build_slidefig_pctfail.py:
% Qwen3-8B = pooled set1-5 base_summ, Gemini true-loop judge (LOOP_ERROR_ANALYSIS_2026-07-10.md);
% Qwen3-14B = ext400/summ_14b, heuristic columns (same file); Gemini = summ@1400 arms, heuristic detector
% (anti_loop_training_log.md 2026-07-21/22).
\begin{tabular}{llrrr}
\toprule
\rowcolor{TableHead}
Model & Compression & Episodes & \shortstack{Failure,\\loop (\%)} & \shortstack{Failure,\\no loop (\%)} \\ \midrule
Qwen3-8B & Summ.\ 4,096 & 1,500 & 84.9 & 67.7 \\
Qwen3-14B & Summ.\ 4,096 & 400 & 70.8 & 47.7 \\
Gemini 3.5 Flash-Lite & Summ.\ 1,400 & 300 & 61.1 & 36.5 \\
Gemini 3.1 Flash & Summ.\ 1,400 & 300 & 51.3 & 23.9 \\
Gemini 3.6 Flash & Summ.\ 1,400 & 300 & 41.4 & 12.8 \\
Gemini 3.1 Pro & Summ.\ 1,400 & 300 & 47.6 & 11.7 \\
\bottomrule
\end{tabular}

\caption{\textbf{Failure rates behind Figure~\ref{fig:loops-hurt}.} For each
model's compressed baseline on CodeGym: the compression setting, the number of
episodes, and the task-failure rate of runs with and without a loop
(Section~\ref{sec:definition}).}
\label{tab:trueloop}
\end{table}

\paragraph{Environment and agent.}
CodeGym~\citep{du2025generalizable} turns static coding problems into
interactive tool-use environments. Each task exposes a function API with
\texttt{Observe()}, task-specific actions, and \texttt{Done()}, and the
environment server scores the agent's answer against the task's gold answer.
Unless noted otherwise, the agent is Qwen3-8B with greedy decoding and a budget
of 30 rounds, where a round is one model turn and a step is one executed
action. Training and evaluation tasks share no task instance or code ID.
Appendix~\ref{app:notes} lists the evaluation runs behind each result.

\paragraph{Loop detector.}
The canonicalization $\kappa$ of Section~\ref{sec:definition} normalizes each
tool call for whitespace and argument order. A step is a looping action when
its canonical call appeared earlier in the episode and the environment returns
a byte-identical observation.
Practical agent frameworks use a similar rule: the OpenHands stuck detector
flags an agent when the same action returns the same observation four or
more times~\citep{openhands}.

\paragraph{Compression.}
When the estimated history exceeds the threshold, the harness keeps the system
prompt, the task, and the six most recent messages, and replaces everything in
between with a summary. The same served model writes the summary at
temperature zero, following an instruction that asks it to keep exact values,
environment returns, and executed actions. The standard threshold is 4,096
tokens; Figures~\ref{fig:compression}a and~\ref{fig:robust}b also use 8,192
and 2,048 tokens and hard truncation at 4,096 tokens. Every comparison holds
the model and the threshold fixed. Table~\ref{tab:battery} lists, for each
context method, the share of episodes in which compression fired.

\subsection{Gemini Runs}
\label{app:frontier}

The Gemini models~\citep{gemini} run the 300 held-out CodeGym tasks in the same harness and
summarize at 1,400 tokens so that compression fires within their shorter
episodes. Figure~\ref{fig:compression}b compares each model's full-history run
with the episodes of its summarization run in which compression fired;
Figure~\ref{fig:loops-hurt} uses all 300 episodes of the summarization run.

\section{Loop Labels}
\label{app:trueloop}

\paragraph{Loop labels in Figures~\ref{fig:loops-hurt} and~\ref{fig:loop-cost}.}
Both figures use the loop rate of Section~\ref{sec:definition}: a run loops
when it contains at least one looping action. Figure~\ref{fig:loops-hurt}
splits each model's compressed-baseline runs by this label
(\cref{tab:trueloop}): 1,500 episodes on five disjoint sets of 300 CodeGym
tasks outside the evaluation set for Qwen3-8B, 400 CodeGym tasks for
Qwen3-14B, and the 300 held-out tasks for the four Gemini models.

\section{How the Policy Uses Its Action Record}
\label{app:probe}

% Floats for this section, placed early so the two-column floats land next to it.
\begin{table}[t]
\centering
\small\setlength{\tabcolsep}{4pt}
% Record-deletion probe (C/A ratio column dropped in review 2026-10-08), compressed baseline vs. our DPO policy (values from
% analysis/probe_assertion/results_{base,nf8k}.json; n=150 points x k=16 completions per variant).
\begin{tabular}{lrrr}
\toprule
\rowcolor{TableHead}
Policy & A: verbatim $\downarrow$ & B: assertion $\downarrow$ & C: deleted $\downarrow$ \\ \midrule
\rowcolor{TableBase} Compressed baseline & 15.3 & 17.6 & 43.3 \\
\rowcolor{TableOurs} DeLoop (ours) & 3.4 & 4.5 & 15.0 \\
\bottomrule
\end{tabular}

\caption{\textbf{Action-record probe.} Re-issue
rate (\%) of the looping action for Qwen3-8B at 150 CodeGym looping points,
with the record verbatim (A), as a one-line assertion (B), or deleted (C);
16 completions per point at temperature 0.7.}
\label{tab:probe-assertion-trained}
\end{table}

\begin{table*}[t]
\centering
\small
\setlength{\tabcolsep}{4pt}
% Held-out policy ranking and frozen-prefix probes (n=250 each), values from analysis/build_oral_tables.py.
% Rows: base; mixdpo (CodeGym + ALFWorld + Open-AgentRL pairs); nf8k (our DPO policy);
% mix3dpo (mixed-data policy); grpo_repeat (looping-action penalty GRPO).
\begin{adjustbox}{max width=\textwidth}
\begin{tabular}{lrrrr}
\toprule
\rowcolor{TableHead}
Policy & \shortstack{$\log p$\\(loop-breaking)} & \shortstack{$\log p$\\(looping)} & Margin & \shortstack{Loop-breaking\\higher (\%)} \\
\midrule
Compressed baseline & $-3.377$ & $-3.514$ & 0.137 & 49.6 \\
Two-source DeLoop-Mix (CodeGym, ALFWorld, Open-AgentRL pairs) & $-3.979$ & $-5.211$ & 1.232 & 69.2 \\
\bottomrule
\end{tabular}
\end{adjustbox}\par

\vspace{4pt}
\begin{adjustbox}{max width=\textwidth}
\begin{tabular}{lrrrrr}
\toprule
\rowcolor{TableHead}
Policy & Looping $\downarrow$ & Loop-breaking & \texttt{Done()} & Other & Unparsed \\
\midrule
Compressed baseline & 15.6 & 20.8 & 3.6 & 45.2 & 14.8 \\
DeLoop (ours) & 6.0 & 19.2 & 24.0 & 33.2 & 17.6 \\
DeLoop-Mix (ours) & 5.2 & 20.4 & 17.2 & 42.4 & 14.8 \\
Two-source DeLoop-Mix (CodeGym, ALFWorld, Open-AgentRL pairs) & 8.4 & 20.4 & 12.4 & 42.4 & 16.4 \\
Looping-action penalty GRPO & 5.6 & 18.8 & 26.0 & 32.0 & 17.6 \\
\bottomrule
\end{tabular}
\end{adjustbox}\par

\caption{\textbf{Action scores and greedy choices after training.} Qwen3-8B at 250 held-out CodeGym loop points. Top:
log-probabilities of the recorded loop-breaking and looping actions, their
margin, and the share of points where the loop-breaking action scores higher.
Bottom: one greedy action from each frozen prefix, by kind (\%).}
\label{tab:policy-mechanism-appendix}
\end{table*}

These probes, which complement \cref{app:ext:probe}, ask how Qwen3-8B uses the
record of an action it has already run.

\paragraph{The record matters, its wording does not.}
We take 150 looping points from training-side tasks that never appear in
evaluation and rebuild the context in five length-matched variants: (A) the
original action--observation block; (B) a one-line assertion of the action and
its result; (C) no record; (D) the assertion inside the summary; and (E) the
returned value without saying that the agent ran the action. We sample 16
completions per point at temperature 0.7 and measure re-issue, the share of
completions that repeat the looping action. For the compressed baseline,
deleting the record raises re-issue from 15.3\% (A) to 43.3\% (C), while
rewording it keeps re-issue close to the verbatim rate (17.6\% for B, 16.5\%
for D, 18.9\% for E).

\paragraph{Training lowers re-issue in every condition.}
On the same points (\cref{tab:probe-assertion-trained}), DeLoop
re-issues 3.4\% of the time with the verbatim record, 4.5\% with the
assertion, and 15.0\% with the record deleted, against 15.3\%, 17.6\%, and
43.3\% for the compressed baseline.

\paragraph{The deletion effect holds on an unselected sample.}
\label{app:population-action-record}
To avoid selecting on loops, we fix 300 decision points after compression from
253 tasks before any model call, stratified by whether the episode eventually
loops. At each point, the untrained policy chooses its next action with the
most recent completed action--observation record present or deleted (eight
samples each). It re-issues that action 3.3\% of the time with the record
present and 48.8\% with it deleted in this stratified sample; weighting the two
strata by their population shares gives a difference of 44.3 points.

\subsection{Training Suppresses the Looping Action}
\label{app:policy-mechanism}

Two probes at 250 loop points from tasks outside all preference-training data
ask what training changes (\cref{tab:policy-mechanism-appendix}). First, we score the recorded loop-breaking action
against the looping action. The untrained policy ranks the loop-breaking
action higher at 49.6\% of points, a coin flip; the two-source DeLoop-Mix variant (CodeGym,
ALFWorld, and Open-AgentRL pairs) raises this to 69.2\%. Second, we freeze each loop prefix
and take one greedy action. Every trained policy re-issues the looping action
less often than the untrained one (15.6\%): 6.0\% for DeLoop, 5.2\%
for DeLoop-Mix, and 5.6\% for looping-action penalty GRPO. The share
that copies the recorded loop-breaking action stays between 18.8\% and 20.8\%
for every policy, while calls to \texttt{Done()} rise from 3.6\% to between
12.4\% and 26.0\%. Training thus suppresses the looping action rather than
teaching one memorized recovery.

\FloatBarrier

\section{Training Details and Additional Results}
\label{app:cure}

This appendix gives the training recipe, the trained baselines of
Table~\ref{tab:main}, and ablations. Unless noted otherwise, every row is
Qwen3-8B on the 300 held-out CodeGym tasks under summarization at 4,096 tokens.

\subsection{Training Details}
\label{app:cure:recipe}

\paragraph{Mining.} We run the compressed agent on CodeGym training tasks and
flag looping actions with the detector of Section~\ref{sec:definition},
comparing canonicalized calls (whitespace and argument order normalized) and
exact observations. Each looping action that the agent later breaks yields one pair: the prompt is the
history just before it, the rejected action is the looping action, and the
chosen action is the first later action in the same rollout that is neither a
looping action nor the same call. Without the second condition, a later repeat
that returns a new observation could be chosen, and both sides of the pair
would be the same text.

\paragraph{No labels.} The miner reads only actions and observations. It never
reads the reward, the task status, or a gold answer, and no LLM judge takes
part. Of the 8,000 CodeGym pairs, 82.6\% come from runs that fail.

\paragraph{Hyperparameters.} The Qwen3-8B DeLoop policy trains LoRA~\citep{hu2022lora} adapters ($r=32$,
$\alpha=64$, all projection modules) from the base model with DPO for one
epoch on 8,000 of the 9,403 mined pairs, with $\beta=0.05$, learning rate
$8\times10^{-6}$, effective batch 16 (batch 1, 16 accumulation steps), and
maximum sequence length 4,500. We chose $\beta$ from
$\{0.02, 0.05, 0.1, 0.2, 0.5\}$ on a separate 128-task screening set, disjoint from the 300 evaluation tasks. Qwen3-14B and Qwen3-32B use the same mining
rule and $\beta$ on 2,331 and 2,244 pairs mined from their own compressed
CodeGym rollouts, and train for two epochs. All policies decode greedily.

\paragraph{Cross-domain pairs.} DeLoop-Mix adds 4,635 cross-domain
pairs to the 8,000 CodeGym pairs (12,635 in total) and keeps the optimizer and
$\beta$. Each pair again contrasts a looping action with a later
loop-breaking action from the same rollout, mined with the same label-free rule:
no reward, task status, gold answer, or other supervision is used. The pairs come from ALFWorld
(1,712), ALFWorld tasks run in the Open-AgentRL harness (1,711), and
Terminal-Bench (1,212). The ALFWorld and Open-AgentRL pairs come from ALFWorld
training games; Table~\ref{tab:envs} evaluates both environments on the 134
unseen validation games, which share no game with the training split. A two-source variant uses the same number of
cross-domain pairs, 2,511 from ALFWorld and 2,124 from the Open-AgentRL
harness.

\subsection{Matched Baselines}
\label{app:cure:grid}

\paragraph{Baselines.} The four offline trained baselines of
Table~\ref{tab:main} share our model, optimizer settings, and budget of 8,000
examples, and differ only in how the examples are built. Outcome DPO uses
outcome-ranked pairs~\citep{song2024eto,dmpo2024} whose rejected side is a random step from a failed run.
SFT on solved traces imitates runs that solve the task~\citep{star2022,rest2023}. Gold-label DPO also
rejects looping actions but uses task outcomes to build its corpus.
Distillation SFT imitates the full-history agent and, like DeLoop, uses no
labels. Gold-action GRPO trains online on the same 8,000 prompts and rewards
an exact match with the gold action.

\paragraph{More data does not close the gap.} At 3,910 and at 8,000 examples,
DeLoop gives the lowest loop rate (18.6\% and 6.1\%) and the fewest steps per
task; the strongest baseline, gold-label DPO, reaches 37.5\% and 22.3\%
(Table~\ref{tab:mechanism-appendix}).

\begin{coltable}[t]
\centering
\small
\setlength{\tabcolsep}{3pt}
% Matched-budget baselines (app:cure:grid). Qwen3-8B, CodeGym-300, summarization at 4,096 tokens.
% Values from the earlier auto-generated dose-matched ladder (rates converted to %); 8,000-example rows match tab:main.
\begin{tabular*}{\linewidth}{@{\extracolsep{\fill}}lcccc@{}}
\toprule
\rowcolor{TableHead}
Method & Examples & \shortstack{Loop\\(\%)\,$\downarrow$} & \shortstack{Success\\(\%)\,$\uparrow$} & \shortstack{Steps\\$\downarrow$} \\ \midrule
Full history & --- & 29.0 & 31.7 & 12.23 \\
\rowcolor{TableBase} Compressed baseline & --- & 57.3 & 25.1 & 15.00 \\
\midrule
Outcome DPO & 3,910 & 49.5 & 25.8 & 12.95 \\
 & 8,000 & 34.7 & 25.7 & 10.99 \\
\addlinespace[2pt]
Gold-label DPO & 3,910 & 37.5 & 26.3 & 10.23 \\
 & 8,000 & 22.3 & 25.7 & 8.10 \\
\addlinespace[2pt]
SFT on solved traces & 3,910 & 44.5 & 23.7 & 11.71 \\
 & 8,000 & 42.7 & 21.3 & 10.61 \\
\addlinespace[2pt]
Distillation SFT & 3,910 & 47.0 & 24.0 & 11.80 \\
 & 8,000 & 43.7 & 26.3 & 11.52 \\
\addlinespace[2pt]
\rowcolor{TableOurs} \textbf{DeLoop (ours)} & 3,910 & 18.6 & 22.2 & 7.47 \\
\rowcolor{TableOurs} & 8,000 & \textbf{6.1} & 25.9 & \textbf{4.67} \\
\bottomrule
\end{tabular*}

\caption{\textbf{Matched baselines at two training budgets.} Qwen3-8B, CodeGym, summarization at 4,096 tokens. Each offline
baseline is trained on 3,910 and on 8,000 examples; the 8,000-example rows
appear in Table~\ref{tab:main}.}
\label{tab:mechanism-appendix}
\end{coltable}

\subsection{More Pairs, Fewer Loops}
\label{app:cure:dose}

The loop rate falls as we add self-mined pairs, from 45.1\% at 1k pairs to
6.1\% at 8k, and steps per task fall from 11.88 to 4.67
(Table~\ref{tab:dose-appendix}). The loop rate levels off after 8k pairs.

\begin{coltable}[t]
\centering
\small
\setlength{\tabcolsep}{3pt}
% Training-data dose (app:cure:dose). Qwen3-8B, CodeGym-300, summarization at 4,096 tokens. Rates in %.
\begin{tabular*}{\linewidth}{@{\extracolsep{\fill}}lccc@{}}
\toprule
\rowcolor{TableHead}
Self-mined pairs & \shortstack{Loop\\(\%)\,$\downarrow$} & \shortstack{Success\\(\%)\,$\uparrow$} & \shortstack{Steps\\$\downarrow$} \\ \midrule
\rowcolor{TableBase} Compressed baseline & 57.3 & 25.1 & 15.00 \\
1k & 45.1 & 26.4 & 11.88 \\
2k & 34.5 & 24.7 & 9.57 \\
3.9k & 18.6 & 22.2 & 7.47 \\
\rowcolor{TableOurs} 8k (DeLoop) & \textbf{6.1} & 25.9 & \textbf{4.67} \\
9.4k & 6.7 & 23.0 & 4.82 \\
\bottomrule
\end{tabular*}

\caption{\textbf{Effect of the number of self-mined pairs.}
Qwen3-8B, CodeGym, summarization at 4,096 tokens; the DeLoop recipe on the number
of self-mined pairs shown.}
\label{tab:dose-appendix}
\end{coltable}

\subsection{How Much Cross-Domain Data to Add}
\label{app:cure:mix}

Every mixture of CodeGym and cross-domain pairs raises task success above the
compressed baseline (25.1\%) and DeLoop (25.9\%), and keeps
the loop rate at or below 24.3\% up to a 50\% cross-domain share
(Table~\ref{tab:recipe-main}). A 37\% share gives the highest success among the
mixtures. At that share, three sources instead of two keep task success
(30.3\% vs.\ 30.9\%) and lower the loop rate from 19.0\% to 10.1\%; this
three-source mixture is DeLoop-Mix.

\begin{coltable}[t]
\centering
\small
\setlength{\tabcolsep}{2.5pt}
% Cross-domain share (app:cure:mix). Qwen3-8B, CodeGym-300, summarization at 4,096 tokens. Rates in %.
\begin{tabular*}{\linewidth}{@{\extracolsep{\fill}}lcccc@{}}
\toprule
\rowcolor{TableHead}
Training data & Share & \shortstack{Loop\\(\%)\,$\downarrow$} & \shortstack{Success\\(\%)\,$\uparrow$} & \shortstack{Steps\\$\downarrow$} \\ \midrule
\rowcolor{TableBase} Compressed baseline & --- & 57.3 & 25.1 & 15.00 \\
CodeGym only (DeLoop) & 0\% & 6.1 & 25.9 & 4.67 \\
+ two sources & 10\% & 8.3 & 27.6 & 5.55 \\
+ two sources & 20\% & 11.9 & 26.8 & 6.74 \\
\rowcolor{TableOurs} + two sources & 37\% & 19.0 & 30.9 & 8.80 \\
\rowcolor{TableOurs} \textbf{+ three sources (DeLoop-Mix)} & 37\% & 10.1 & 30.3 & 5.94 \\
+ two sources & 50\% & 24.3 & 27.3 & 9.48 \\
Cross-domain only & 100\% & 55.3 & 31.3 & 15.13 \\
\bottomrule
\end{tabular*}

\caption{\textbf{Cross-domain data mixtures.}
Qwen3-8B, CodeGym, summarization at 4,096 tokens; DPO on the 8,000 CodeGym
pairs plus cross-domain pairs at the share shown (last row: cross-domain pairs
only). The highlighted rows hold the share at 37\%; the three-source row is
DeLoop-Mix in Table~\ref{tab:main}.}
\label{tab:recipe-main}
\end{coltable}

\subsection{Comparison with Agent-R-Style Splicing}
\label{app:cure:lowdose-ports}

Agent-R~\citep{agentr} learns from whole trajectories and task outcomes: it
splices a failed path onto a reward-selected branch at the point where the
two diverge. We adapt this divergence splicing to our rollouts, where it yields
1,000 training pairs, and run DeLoop on 1,000 self-mined pairs next to it.
Without outcome labels, self-mined pairs lower the loop rate more (45.1\%
vs.\ 52.3\%), take fewer steps per task (11.88 vs.\ 13.62), and keep higher
task success (26.4\% vs.\ 24.0\%; Table~\ref{tab:lowdose-ports}).

\begin{table*}[t]
\centering
\small
\setlength{\tabcolsep}{4pt}
% Low-budget trajectory-level adaptations (app:cure:lowdose-ports). Qwen3-8B, CodeGym-300, summarization at
% 4,096 tokens. Rates in %; tokens per task in thousands. Review 2026-10-08: the 383-pair block
% (ours 56.3/26.0/14.14/23.2; ETO-style 55.3/26.7/14.77/23.1; DEPO-style 62.7/29.0/15.05/23.3) moved to Limitations.
\begin{tabular*}{\textwidth}{@{\extracolsep{\fill}}lcccccc@{}}
\toprule
\rowcolor{TableHead}
Method & Budget & Labels & Loop (\%) $\downarrow$ & Success (\%) $\uparrow$ & Steps $\downarrow$ & Tokens/task (k) $\downarrow$ \\ \midrule
\rowcolor{TableBase} Compressed baseline & --- & none & 57.3 & 25.1 & 15.00 & 23.9 \\
\midrule
\rowcolor{TableOurs} DeLoop (ours) & 1,000 & none & \textbf{45.1} & 26.4 & \textbf{11.88} & \textbf{22.4} \\
Agent-R-style divergence splice & 1,000 & task success & 52.3 & 24.0 & 13.62 & 27.1 \\
\bottomrule
\end{tabular*}

\caption{\textbf{Comparison with Agent-R-style splicing.} Qwen3-8B, CodeGym, summarization at
4,096 tokens. Both trained rows use 1,000 training pairs. Labels: the
supervision each method needs.}
\label{tab:lowdose-ports}
\end{table*}

\subsection{Hyperparameters, Mining Rules, and Model Size}
\label{app:cure:hparam}
\label{app:cure:miner-signal-scale}

\paragraph{Hyperparameters.} Changing one field of the recipe at a time, every
setting lowers the loop rate below the compressed baseline's 57.3\%
(Table~\ref{tab:hparam-ablation-appendix}). The default recipe keeps the
highest task success (25.9\%), and $\beta=0.05$ gives the lowest loop rate of
the $\beta$ values. A higher learning rate, a larger LoRA rank, or a second
epoch on the 8,000 pairs lowers the loop rate further, to 0.7--1.7\%, with
1.60--3.59 steps per task.

\paragraph{Mining rules.} Changing one mining rule at a time, re-mining the
same rollouts, and training on 8,000 pairs keeps the loop rate at or below
7.0\%; the default rule keeps the highest task success
(Table~\ref{tab:miner-ablation-appendix}). The same table gives activity for
the SFT control of Table~\ref{tab:ablation}: it shortens episodes
less than the pair (11.71 rounds per task, against 5.47 for full pairs and
16.34 for the compressed baseline).

\paragraph{Model size.} Pairs transfer across model sizes
(Table~\ref{tab:pair-transfer-appendix}). Qwen3-8B trained on 2,331 pairs mined
from Qwen3-14B lowers its loop rate from 57.3\% to 35.0\% and raises task
success from 25.1\% to 28.7\%. Qwen3-14B trained on the 8,000 Qwen3-8B pairs
lowers its loop rate from 42.9\% to 9.3\%.

\begin{coltable}[t]
\centering
\small
\setlength{\tabcolsep}{3pt}
% Hyperparameter, exposure and dose ablations (app:cure:hparam). Qwen3-8B, CodeGym-300, summarization at 4,096
% tokens. Rates in %. One field changed per row.
\begin{tabular*}{\linewidth}{@{\extracolsep{\fill}}lccc@{}}
\toprule
\rowcolor{TableHead}
Setting & \shortstack{Loop\\(\%)\,$\downarrow$} & \shortstack{Success\\(\%)\,$\uparrow$} & \shortstack{Steps\\$\downarrow$} \\ \midrule
\rowcolor{TableBase} Compressed baseline & 57.3 & 25.1 & 15.00 \\
\rowcolor{TableOurs} Default recipe (DeLoop) & 6.1 & \textbf{25.9} & 4.67 \\
\midrule
\rowcolor{TableGroup} \multicolumn{4}{l}{\textit{DPO $\beta$}} \\
$\beta=0.01$ & 17.3 & 16.3 & 2.06 \\
$\beta=0.1$ & 17.7 & 23.0 & 6.88 \\
$\beta=0.2$ & 36.7 & 25.0 & 10.57 \\
\rowcolor{TableGroup} \multicolumn{4}{l}{\textit{Learning rate}} \\
lr $2\times10^{-6}$ & 35.3 & 23.7 & 10.33 \\
lr $3\times10^{-5}$ & 0.7 & 19.3 & 1.60 \\
\rowcolor{TableGroup} \multicolumn{4}{l}{\textit{LoRA rank}} \\
$r=8$ ($\alpha=16$) & 25.3 & 24.0 & 7.84 \\
$r=64$ ($\alpha=128$) & 1.3 & 24.0 & 3.59 \\
\rowcolor{TableGroup} \multicolumn{4}{l}{\textit{Epochs and pairs}} \\
8,000 pairs, 2 epochs & 1.7 & 25.3 & 2.85 \\
9,403 pairs, 2 epochs & 7.0 & 20.7 & 2.80 \\
500 pairs, 1 epoch & 52.3 & 19.7 & 14.08 \\
\bottomrule
\end{tabular*}

\caption{\textbf{Training hyperparameter ablations.} Qwen3-8B, CodeGym, summarization at
4,096 tokens. Each block changes one field of the default recipe.}
\label{tab:hparam-ablation-appendix}
\end{coltable}

\begin{coltable}[t]
\centering
\small
\setlength{\tabcolsep}{2.5pt}
% Mining-rule and pair-signal variants (app:cure:hparam). Qwen3-8B, CodeGym-300, summarization at 4,096 tokens.
% Rates in %. Rounds = agent turns per task. Pair-signal rows were tab_signal_decomposition_appendix.
\begin{tabular*}{\linewidth}{@{\extracolsep{\fill}}lcccc@{}}
\toprule
\rowcolor{TableHead}
Setting & Mined & \shortstack{Loop\\(\%)\,$\downarrow$} & \shortstack{Success\\(\%)\,$\uparrow$} & \shortstack{Rounds\\$\downarrow$} \\ \midrule
\rowcolor{TableBase} Compressed baseline & --- & 57.3 & 25.1 & 16.34 \\
\rowcolor{TableOurs} Default (DeLoop) & 9,403 & 6.1 & \textbf{25.9} & 5.47 \\
SFT (chosen action only) & --- & 37.7 & 21.3 & 11.71 \\
\rowcolor{TableGroup} \multicolumn{5}{l}{\textit{Mining rule (one change each)}} \\
Random later chosen & 9,357 & 3.7 & 23.3 & 4.61 \\
Exact-string match & 9,294 & 7.0 & 22.7 & 5.02 \\
Near-identical obs. & 10,265 & 3.0 & 20.0 & 3.85 \\
\bottomrule
\end{tabular*}

\caption{\textbf{Mining-rule and chosen-only ablations.} Qwen3-8B, CodeGym, summarization at 4,096 tokens;
8,000 training pairs per row. ``Mined'': pairs the rule yields. The variants
pick a random later loop-breaking action, match actions as exact strings, or
treat observations with similarity ratio $\geq 0.9$ as the same. SFT trains
on the loop-breaking (chosen) action alone. Rounds: agent turns per task.}
\label{tab:miner-ablation-appendix}
\label{tab:signal-decomposition-appendix}
\end{coltable}

\begin{coltable}[t]
\centering
\small
\setlength{\tabcolsep}{2.5pt}
% Cross-size pair transfer (app:cure:hparam). CodeGym-300, summarization at 4,096 tokens. Rates in %.
% Rounds = agent turns per task. The Qwen3-14B "ours" row is the main-text policy (macros \fourteenBHighBetaEpTwo*).
\begin{tabular*}{\linewidth}{@{\extracolsep{\fill}}llccc@{}}
\toprule
\rowcolor{TableHead}
Model & Training pairs & \shortstack{Loop\\(\%)\,$\downarrow$} & \shortstack{Success\\(\%)\,$\uparrow$} & \shortstack{Rounds\\$\downarrow$} \\ \midrule
\rowcolor{TableBase} Qwen3-8B & none & 57.3 & 25.1 & 16.34 \\
\rowcolor{TableOurs} Qwen3-8B & 8,000 own & 6.1 & 25.9 & 5.47 \\
Qwen3-8B & 2,331 from 14B & 35.0 & 28.7 & 11.28 \\
\midrule
\rowcolor{TableBase} Qwen3-14B & none & 42.9 & 45.1 & 15.57 \\
\rowcolor{TableOurs} Qwen3-14B & 2,331 own & 12.3 & 52.3 & 8.26 \\
Qwen3-14B & 8,000 from 8B & 9.3 & 44.3 & 7.21 \\
\bottomrule
\end{tabular*}

\caption{\textbf{Preference-pair transfer across model sizes.}
CodeGym, summarization at 4,096 tokens; DPO on pairs mined from the model's own
rollouts or the other model's. Highlighted rows: DeLoop from
Figure~\ref{fig:scale}. Rounds: agent turns per task.}
\label{tab:pair-transfer-appendix}
\end{coltable}
\FloatBarrier

% appendix_cure_b.tex (rewritten 2026-10-08, appendix pass E): middle part of the training appendix.
% Continues the section opened in appendix_cure_a.tex. Caveats from the earlier version now live in
% Limitations; numbers follow Tables 1-2 of the main text.
\subsection{Per-Step Selectivity}
\label{app:cure:selectivity}

Not every repeated call is wasted: re-checking a state the agent has changed
returns a new observation. We split repeated calls into looping actions, which
get back the same observation, and informative repeats, which get back a new
one. Selectivity is the share of repeats that are informative, pooled over all
scored steps.

Training removes looping actions far more than informative repeats
(\cref{tab:selectivity-appendix}). DeLoop lowers looping actions from
24.9\% to 2.5\% of steps and informative repeats from 10.6\% to 5.8\%, so
selectivity rises from 29.9\% to 69.8\%; DeLoop-Mix reaches
74.2\%. The trained policies still re-check the environment, but they rarely
re-issue a call whose answer they already have.

\begin{table}[h!]
\centering
\small
\setlength{\tabcolsep}{4pt}
% Per-step selectivity (rewritten 2026-10-08, appendix pass E). Qwen3-8B, CodeGym-300, summarization at 4,096
% tokens, pooled over 300 tasks x 3 evaluation runs. Source values (fractions of scored steps) from the earlier
% auto-generated table: base 0.2487 / 0.1060 / 0.299 / 13504; our DPO 0.0250 / 0.0578 / 0.698 / 4204;
% mixed-data 0.0260 / 0.0748 / 0.742 / 5347. Shown as percentages with one decimal.
\begin{tabular}{lcccc}
\toprule
\rowcolor{TableHead}
Policy & Looping actions (\% of steps) $\downarrow$ & Informative repeats (\% of steps) & Selectivity (\%) $\uparrow$ & Scored steps \\ \midrule
\rowcolor{TableBase} Compressed baseline & 24.9 & 10.6 & 29.9 & 13,504 \\
\rowcolor{TableOurs} DeLoop (ours) & 2.5 & 5.8 & 69.8 & 4,204 \\
\rowcolor{TableOurs} DeLoop-Mix (ours) & 2.6 & 7.5 & 74.2 & 5,347 \\
\bottomrule
\end{tabular}

\caption{\textbf{Informative repeats and looping actions.} Qwen3-8B on CodeGym under summarization at 4,096 tokens, pooled
over the 300 tasks and three evaluation runs. Selectivity is the share of
repeats that return new information.}
\label{tab:selectivity-appendix}
\end{table}
\FloatBarrier

\subsection{Training Objectives on the Same Loop Signal}
\label{app:cure:objective-family}

DPO, KTO, and IPO in Table~\ref{tab:ablation} and
Figure~\ref{fig:objectives} train Qwen3-8B on the same 8,000 self-mined
CodeGym pairs; DPO uses the DeLoop recipe (\cref{app:cure:recipe}), and KTO and IPO
use their published default settings. IPO~\citep{azar2024ipo}
replaces the log-sigmoid of DPO with a squared loss that bounds the
preference margin. KTO~\citep{ethayarajh2024kto} drops the pairing and scores
each action on its own, with loop-breaking actions as desirable and looping
actions as undesirable; it still cuts the loop rate to 12.6\%. Looping-action
penalty GRPO uses no pairs; it penalizes every flagged step in fresh online
rollouts (\cref{app:cure:grpo-diagnosis}). All four objectives shorten
episodes: steps per task fall from 15.00 to between 4.67 (DPO) and 6.72
(IPO), and tokens per task from 23.9k to between 12.4k and 13.6k.

\subsection{Inference-Time Controls}
\label{app:cure:inference}

\Cref{tab:inference-controls} lists every training-free control we ran on the
compressed baseline. The exact duplicate blocker refuses to run a call that
repeats an earlier one, so its loop rate counts attempted repeats.
Temperature 0.7 and a token repetition penalty~\citep{ctrl2019} of 1.2 change decoding.
Reflexion, stuck-stop, and stuck-nudge act when a stuck detector fires: the
model critiques itself before acting, the episode ends, or the model gets a
reminder.

Inference-time controls do not break the loop. Every control leaves the loop
rate at 46.7\% or above, against 6.1\% for DeLoop. The blocker comes
closest, but steps per task barely move (14.71 against 15.00) because the
agent keeps proposing the blocked call. The decoding controls raise the loop
rate to 63.3\% and 65.7\%, and the within-episode controls stay between
55.3\% and 58.7\%.

\begin{table}[h!]
\centering
\small
\setlength{\tabcolsep}{14pt}
% Inference-time controls (rewritten 2026-10-08, appendix pass E). Qwen3-8B, CodeGym-300, summarization at 4,096
% tokens. Names and numbers match Table 1 (tables/v2/tab_main.tex) for the shared rows. Tokens/task (k) totals:
% baseline 23,915; blocker 1,312+23,549; temperature 1,100+17,438; repetition penalty 1,098+13,587 (macros.tex
% L1732-1746); Reflexion 24.1 and ours 12.7 (macros_gold.tex). Stuck-stop and stuck-nudge are sums of the rounded
% input/output values of the earlier table (1.1k+20.5k, 1.1k+22.1k).
\begin{tabular}{lcccc}
\toprule
\rowcolor{TableHead}
Control & Loop (\%) $\downarrow$ & Success (\%) $\uparrow$ & Steps $\downarrow$ & Tokens/task (k) $\downarrow$ \\ \midrule
\rowcolor{TableBase} Compressed baseline & 57.3 & 25.1 & 15.00 & 23.9 \\
\rowcolor{TableGroup}\multicolumn{5}{l}{\textit{Blocking and decoding}} \\
Exact duplicate blocker & 46.7 & 25.3 & 14.71 & 24.9 \\
Temperature 0.7 & 63.3 & 29.3 & 16.99 & 18.5 \\
Repetition penalty 1.2 & 65.7 & 29.0 & 16.85 & 14.7 \\
\rowcolor{TableGroup}\multicolumn{5}{l}{\textit{Within-episode controls (act when a stuck detector fires)}} \\
Reflexion on stuck & 55.3 & 24.3 & 14.62 & 24.1 \\
Stuck-stop (end the episode) & 57.3 & 21.7 & 13.08 & 21.6 \\
Stuck-nudge (remind the model) & 58.7 & 23.7 & 14.85 & 23.2 \\
\rowcolor{TableGroup}\multicolumn{5}{l}{\textit{Ours}} \\
\rowcolor{TableOurs} DeLoop (ours) & 6.1 & 25.9 & 4.67 & 12.7 \\
\bottomrule
\end{tabular}

\caption{\textbf{Inference-time controls.} Qwen3-8B on CodeGym under summarization at 4,096 tokens. Every control
runs on the compressed baseline without training.}
\label{tab:inference-controls}
\end{table}
\FloatBarrier

\subsection{GRPO Variants}
\label{app:cure:grpo-diagnosis}

We train Qwen3-8B with online GRPO~\citep{shao2024deepseekmath} under summarization at 4,096 tokens and
vary the reward, sampling, or prompts (\cref{tab:grpo-diagnosis-appendix}).
Looping-action penalty GRPO penalizes every looping action; one version also
mixes in prompts from other environments. Episode-level loop reward GRPO gives
one sparse reward per episode, and loop-weighted SFT replaces RL with weighted
imitation. Gold-action GRPO, the labeled baseline of Table~\ref{tab:main},
rewards an exact match with a gold action; its variants add partial credit and
more diverse sampling. A group is informative when its rewards differ, so that
it yields a gradient.

\paragraph{The per-step penalty is the strongest signal.} It cuts the loop
rate to 15.0\%, and to 18.3\% with 28.3\% task success when prompts are mixed.
A sparse episode-level reward leaves 47.3\% of runs looping, and loop-weighted
SFT 42.3\%.

\paragraph{Gold-action rewards give GRPO little to learn from.} With exact
matching, only 2.7\% of groups are informative and the loop rate stays at
52.7\%. Partial credit raises informative groups to 5.3\% and diverse sampling
to 20.5\%, and the loop rate falls with them, to 49.3\% and 32.7\%.

\paragraph{The penalty holds across context settings.} Trained once under
summarization at 4,096 tokens, the policy loops in 13.7\% of tasks with the
full history and in 11.3\%, 15.0\%, and 16.0\% under summarization at 8,192,
4,096, and 2,048 tokens, against 29.0\%, 51.0\%, 57.3\%, and 64.7\% for the
untrained agent in the same settings. Under truncation, which it never saw in
training, it loops in 24.7\% of tasks against 65.3\%. Task success rises from
20.0\% to 24.3\% at 2,048 tokens and from 18.0\% to 25.7\% under
truncation.

\begin{table*}[htbp]
\centering
\small
\setlength{\tabcolsep}{3pt}
% GRPO variants (rewritten 2026-10-08, appendix pass E). Qwen3-8B, CodeGym-300, summarization at 4,096 tokens.
% Structural loop rate, success, steps, tokens (23,915 -> 23.9k etc.) and informative-group shares from the earlier
% auto-generated table; judged-loop column dropped. Names follow Tables 1-2 of the main text.
\begin{tabular}{llccccc}
\toprule
\rowcolor{TableHead}
Method & Reward or signal & Loop (\%) $\downarrow$ & Success (\%) $\uparrow$ & Steps $\downarrow$ & \shortstack{Tokens/\\task (k) $\downarrow$} & \shortstack{Informative\\groups (\%)} \\ \midrule
\rowcolor{TableBase} Compressed baseline & none & 57.3 & 25.1 & 15.00 & 23.9 & --- \\
\rowcolor{TableGroup}\multicolumn{7}{l}{\textit{Loop signal (no labels)}} \\
\rowcolor{TableOurs} Looping-action penalty GRPO & per-step penalty & 15.0 & 24.7 & 6.39 & 13.4 & --- \\
\quad + mixed prompts & per-step penalty & 18.3 & 28.3 & 7.08 & 13.8 & --- \\
Episode-level loop reward GRPO & one reward per episode & 47.3 & 25.3 & 13.53 & 21.8 & --- \\
Loop-weighted SFT & weighted imitation & 42.3 & 19.0 & 9.92 & 28.6 & --- \\
\rowcolor{TableGroup}\multicolumn{7}{l}{\textit{Gold-action reward (labels)}} \\
Gold-action GRPO & exact match & 52.7 & 25.0 & 14.23 & 21.0 & 2.7 \\
\quad + partial credit & partial match & 49.3 & 27.0 & 13.22 & 22.2 & 5.3 \\
\quad + partial credit, diverse sampling & partial match & 32.7 & 27.3 & 11.08 & 20.2 & 20.5 \\
\bottomrule
\end{tabular}

\caption{\textbf{GRPO reward and sampling variants.} Qwen3-8B on CodeGym under summarization at 4,096 tokens. Informative
groups are the share of sampled groups whose rewards differ.}
\label{tab:grpo-diagnosis-appendix}
\end{table*}
\FloatBarrier

\subsection{A Stricter Chosen Rule Does Not Help}
\label{app:cure:chosen-rule}

Requiring the agent to keep receiving new observations after the
loop-breaking action does not improve the pairs. At 1,726 pairs from the same
rollouts, this stricter rule gives a loop rate of 41.7\% and task success of
23.0\%, against 38.7\% and 25.0\% for the standard rule at the same number of
pairs.

\subsection{Combining SFT with DPO}
\label{app:cure:sft-dpo}

When SFT and DPO are staged, DPO must come last
(\cref{tab:sft-dpo-appendix}). In a separate screening evaluation, SFT on
solved CodeGym traces after DPO reopens loops (6.5\% to 29.7\%), and a final
DPO stage closes them again (2.3\%).

\begin{table*}[t]
\centering
\small
\setlength{\tabcolsep}{7.5pt}
% Sequential SFT/DPO recipes (rewritten 2026-10-08, appendix pass E). Qwen3-8B, CodeGym, summarization at 4,096
% tokens. Upper block: main 300-task evaluation; lower block: separate screening evaluation of objective order.
% Values from the earlier auto-generated table (fractions shown as percentages).
\begin{tabular}{l>{\raggedright\arraybackslash}p{6.2cm}ccc}
\toprule
\rowcolor{TableHead}
Training sequence & SFT demonstrations & Loop (\%) $\downarrow$ & Success (\%) $\uparrow$ & Steps $\downarrow$ \\ \midrule
\rowcolor{TableGroup}\multicolumn{5}{l}{\textit{Main evaluation}} \\
\rowcolor{TableOurs} DPO only (DeLoop) & --- & 6.1 & 25.9 & 4.67 \\
% SFT $\rightarrow$ DPO & successful CodeGym, ALFWorld, and Open-AgentRL trajectories & 19.0 & 30.7 & 8.37 \\ % removed 2026-10-11 (success-selected ALFWorld data)
\rowcolor{TableGroup}\multicolumn{5}{l}{\textit{Screening evaluation: objective order}} \\
DPO only & --- & 6.5 & 16.4 & 4.74 \\
DPO $\rightarrow$ SFT & solved CodeGym traces & 29.7 & 12.5 & 10.79 \\
DPO $\rightarrow$ SFT $\rightarrow$ DPO & solved CodeGym traces & 2.3 & 21.1 & 4.62 \\
\bottomrule
\end{tabular}

\caption{\textbf{Sequential SFT and DPO training.} Qwen3-8B
on CodeGym under summarization at 4,096 tokens. The lower block is a separate
screening evaluation, compared only within the block.}
\label{tab:sft-dpo-appendix}
\end{table*}
\FloatBarrier

% appendix_cure_c.tex -- last part of the repair appendix (rewritten 2026-10-08 in the main-text style).
\subsection{A Likelihood Anchor Brings Loops Back}
\label{app:cure:rpo}

Plain DPO suits self-mined pairs better than DPO with a likelihood anchor.
RPO adds a negative log-likelihood term on the chosen action to the DPO loss,
with weight $\alpha=1.0$. On the same 8,000 self-mined CodeGym pairs, the
anchor raises the loop rate of Qwen3-8B from 6.1\% to 27.0\%, steps per task
from 4.67 to 7.90, and tokens per task from 12.7k to 24.8k, and task
success goes from 25.9\% to 21.3\%. The anchor raises loops on other training
data as well: from 19.0\% to 42.1\% on a two-source mixed corpus, and from
34.7\% to 37.3\% on the outcome-ranked pairs of Table~\ref{tab:main}.

The training logs show what the anchor changes. Plain DPO pushes the looping
action far down and lowers the loop-breaking action only slightly; with the
anchor, the loop-breaking action rises toward imitation.
Appendix~\ref{app:theory:anchor} derives this trade-off from the gradient of
the objective and gives the log-probabilities. Imitation alone leaves many
loops in place (37.7\% for the chosen-only row of Table~\ref{tab:ablation}).

\subsection{Cross-Domain Pairs from Public Trajectories}
\label{app:cure:mix-mechanism}

The gain from cross-domain pairs does not depend on one trajectory
collection. We keep the self-mined CodeGym pairs and the size of the
cross-domain portion fixed, build the cross-domain pairs from a public
collection instead, either SWE-agent~\citep{swe_agent} or OpenHands~\citep{openhands} coding trajectories or
WebArena~\citep{webarena} browser trajectories, and train Qwen3-8B with the same DPO recipe
(\cref{tab:mix-source-expansion-appendix}). With every source, the CodeGym
loop rate falls from 57.3\% to between 17.3\% and 22.7\%, task success rises
from 25.1\% to between 26.0\% and 28.7\%, and steps per task fall from 15.00
to 8.00 or fewer. The same policies also loop less on ALFWorld (33.6\% to
46.3\%, against 55.2\%), and the WebArena pairs raise ALFWorld task success
from 37.3\% to 41.8\%.

\begin{coltable}[t]
\centering
\footnotesize
\setlength{\tabcolsep}{2.2pt}
\resizebox{\linewidth}{!}{% Cross-domain source substitution (rewritten 2026-10-08, agent F). Qwen3-8B; CodeGym and ALFWorld columns
% from the earlier source-expansion table (values unchanged, rates converted to percentages).
\begin{tabular}{@{}lccccc@{}}
\toprule
 & \multicolumn{3}{c}{CodeGym} & \multicolumn{2}{c}{ALFWorld} \\
\cmidrule(lr){2-4}\cmidrule(lr){5-6}
Cross-domain source & Loop (\%) $\downarrow$ & Success (\%) $\uparrow$ & Steps $\downarrow$ & Loop (\%) $\downarrow$ & Success (\%) $\uparrow$ \\ \midrule
\rowcolor{TableBase} Compressed baseline & 57.3 & 25.1 & 15.00 & 55.2 & 37.3 \\
\rowcolor{TableOurs} SWE-agent & 17.3 & 26.0 & 7.18 & 35.8 & 36.6 \\
\rowcolor{TableOurs} OpenHands & 21.0 & 28.7 & 7.88 & 33.6 & 37.3 \\
\rowcolor{TableOurs} WebArena & 22.7 & 28.3 & 8.00 & 46.3 & 41.8 \\
\bottomrule
\end{tabular}
}
\caption{\textbf{Cross-domain pairs from public trajectories.} Qwen3-8B on CodeGym and ALFWorld with summarization. Each shaded row
trains on the self-mined CodeGym pairs plus cross-domain pairs from the named
public source.}
\label{tab:mix-source-expansion-appendix}
\end{coltable}

\subsection{General Capability Is Preserved}
\label{app:cure:livecodebench}

Training on loops leaves general ability intact. We evaluate the untrained
Qwen3-8B, DeLoop, and DeLoop-Mix on three non-agentic
benchmarks outside agent tasks (\cref{tab:livecodebench}): the 182
LiveCodeBench~\citep{livecodebench} problems released from 2025 onward, with the official prompts
and greedy pass@1; MT-Bench~\citep{mtbench}, with 80 two-turn questions, the official category
temperatures, and one LLM judge for all three models; and IFEval~\citep{ifeval}, with 541
prompts and the official strict-prompt scorer. On LiveCodeBench, the base
model solves 53 problems, DeLoop 52, and DeLoop-Mix 54. DeLoop
scores higher than the base model on MT-Bench (4.463 against
4.225) and on IFEval (83.5\% against 82.8\%), and DeLoop-Mix
reaches 83.4\% on IFEval.

\begin{coltable}[t]
\centering
\footnotesize
\setlength{\tabcolsep}{3pt}
\resizebox{\linewidth}{!}{% General capability checks (rewritten 2026-10-08, agent F): LiveCodeBench, MT-Bench and IFEval merged into one
% table. Values from the earlier tab_livecodebench, tab_mtbench_control (mean column) and tab_ifeval_control.
\begin{tabular}{@{}lccc@{}}
\toprule
Policy & LiveCodeBench & MT-Bench & IFEval (\%) \\ \midrule
\rowcolor{TableBase} Qwen3-8B (untrained) & 53/182 & 4.225 & 82.8 \\
\rowcolor{TableOurs} DeLoop (ours) & 52/182 & 4.463 & 83.5 \\
\rowcolor{TableOurs} DeLoop-Mix (ours) & 54/182 & 4.213 & 83.4 \\
\bottomrule
\end{tabular}
}
\caption{\textbf{General capability evaluation.} Qwen3-8B on LiveCodeBench (problems solved of
182, greedy pass@1), MT-Bench (mean judge score, 1--10), and IFEval
(strict-prompt accuracy).}
\label{tab:livecodebench}
\end{coltable}

\subsection{Repair Across Memory Policies}
\label{app:compressor-repair}

\Cref{tab:compressor-repair-main} gives the numbers behind
Figure~\ref{fig:robust}c. The seven memory policies are retrieval memory, an
instruction ledger, Claude-style compaction, a LangChain summary, and
summaries augmented with an action trace, an action warning, or an action
ledger. Under every one of them, DeLoop-Mix lowers the loop rate,
from 42.7\%--62.7\% to 6.3\%--20.7\%, along with steps per task, and it
cuts tokens per task by more than half. Task
success rises under four of the seven, by up to 5.7 points with the action
ledger.

\begin{table*}[htbp]
\centering
\small
\setlength{\tabcolsep}{3pt}
% Memory-policy repair table (rewritten 2026-10-08, agent F): full numbers behind Figure 5c. Values unchanged;
% rates converted to percentages; the realized-compression column moved to Limitations.
% 2026-10-12: input/output token columns replaced by total tokens/task (author: report totals only until the split is verified).
% Totals = tools/efficiency_metrics.py mean_est_tokens_per_task (base rows match tab_battery; DeLoop-Mix rows from the zip04 runs).
\begin{tabular*}{\textwidth}{@{\extracolsep{\fill}}llcccc@{}}
\toprule
\rowcolor{TableHead}
Memory policy & Policy & Loop (\%) $\downarrow$ & Success (\%) $\uparrow$ & Steps $\downarrow$ & \shortstack{Tokens/\\task (k) $\downarrow$} \\ \midrule
Retrieval memory & Compressed baseline & 42.7 & 27.7 & 13.34 & 26.1 \\
\rowcolor{TableOurs}  & DeLoop-Mix (ours) & \textbf{10.7} & 31.0 & 6.93 & 11.8 \\
\addlinespace[2pt]
Instruction ledger & Compressed baseline & 50.0 & 27.0 & 12.40 & 19.5 \\
\rowcolor{TableOurs}  & DeLoop-Mix (ours) & \textbf{7.0} & 25.3 & 5.16 & 8.8 \\
\addlinespace[2pt]
Claude-style compact & Compressed baseline & 50.3 & 26.3 & 13.67 & 20.9 \\
\rowcolor{TableOurs}  & DeLoop-Mix (ours) & \textbf{6.3} & 25.7 & 5.84 & 9.3 \\
\addlinespace[2pt]
LangChain summary & Compressed baseline & 51.7 & 26.3 & 13.25 & 20.7 \\
\rowcolor{TableOurs}  & DeLoop-Mix (ours) & \textbf{8.7} & 26.3 & 6.01 & 9.9 \\
\addlinespace[2pt]
Action trace & Compressed baseline & 52.0 & 25.0 & 14.21 & 30.9 \\
\rowcolor{TableOurs}  & DeLoop-Mix (ours) & \textbf{20.7} & 27.7 & 7.67 & 14.2 \\
\addlinespace[2pt]
Action warning & Compressed baseline & 54.0 & 24.0 & 14.12 & 22.9 \\
\rowcolor{TableOurs}  & DeLoop-Mix (ours) & \textbf{9.3} & 27.3 & 5.99 & 9.7 \\
\addlinespace[2pt]
Action ledger & Compressed baseline & 62.7 & 23.3 & 15.51 & 24.5 \\
\rowcolor{TableOurs}  & DeLoop-Mix (ours) & \textbf{11.3} & 29.0 & 5.60 & 9.6 \\
\bottomrule
\end{tabular*}

\caption{\textbf{DeLoop-Mix across memory policies.} Qwen3-8B on CodeGym. Each memory policy has a compressed-baseline row
and a DeLoop-Mix row (shaded) on the same tasks. Tokens are estimated
tokens per task (Appendix~\ref{app:notes}).}
\label{tab:compressor-repair-main}
\end{table*}

\subsection{Other Model Families and Sizes}
\label{app:cure:crossmodel}

\paragraph{Llama-3.1-8B.} Llama~\citep{llama3} loops natively. With the full history,
Llama-3.1-8B loops in 44.0\% of CodeGym episodes, and under summarization at
4,096 tokens it loops in 43.7\% (\cref{tab:crossmodel}). We train it with DPO
on pairs mined from its own rollouts with the same rule. The trained policy
loops in 2.0\% of episodes, against 43.7\% for the compressed baseline,
and steps per task fall from 17.51 to 5.72, while task success goes from
17.8\% to 19.4\%.

\paragraph{Qwen3-14B and Qwen3-32B.} Compression raises looping at both
sizes and costs task success. Summarization at 4,096 tokens raises the loop
rate from 33.3\% to 42.9\% on Qwen3-14B and from 26.0\% to 38.5\% on
Qwen3-32B, and lowers task success by 8.9 and 9.0 points. We train each model on pairs
mined from its own compressed rollouts with the recipe of
\cref{app:cure:recipe}. Both policies loop less than the agent with the full
history (12.3\% and 12.9\%), and they win back most of the success that
compression costs: 7.2 of the 8.9 points at 14B and 7.1 of the 9.0 points at
32B (\cref{tab:crossmodel}).

DeLoop at 32B also helps when the history is not compressed. Evaluated with
the full history, it lowers the loop rate from 25.0\% to 11.7\% and raises
task success from 46.3\% to 51.3\%, compared with a base-model run matched to
this test.

\begin{coltable}[t]
\centering
\footnotesize
\setlength{\tabcolsep}{2.5pt}
\resizebox{\linewidth}{!}{%
\begin{tabular}{@{}llccc@{}}
\toprule
Model & Setting & Loop (\%) $\downarrow$ & Success (\%) $\uparrow$ & Steps $\downarrow$ \\ \midrule
Llama-3.1-8B & Full history & 44.0 & 17.5 & 17.41 \\
 & Compressed baseline & 43.7 & 17.8 & 17.51 \\
\rowcolor{TableOurs} & DeLoop (ours) & 2.0 & 19.4 & 5.72 \\
\addlinespace[2pt]
Qwen3-14B & Full history & 33.3 & 54.0 & 14.36 \\
 & Compressed baseline & 42.9 & 45.1 & 15.28 \\
\rowcolor{TableOurs} & DeLoop (ours) & 12.3 & 52.3 & 8.18 \\
\addlinespace[2pt]
Qwen3-32B & Full history & 26.0 & 46.7 & 12.97 \\
 & Compressed baseline & 38.5 & 37.7 & 13.54 \\
\rowcolor{TableOurs} & DeLoop (ours) & 12.9 & 44.8 & 8.82 \\
\bottomrule
\end{tabular}}
\caption{\textbf{Other model families and sizes.} CodeGym. For each model: the agent with the full history, the
compressed baseline (summarization at 4,096 tokens), and DeLoop,
trained on that model's self-mined pairs and evaluated under summarization at
4,096 tokens.}
\label{tab:crossmodel}
\end{coltable}
\FloatBarrier

% appendix_theory.tex -- formal account of loops and their repair.
% Rewritten 2026-10-08 (appendix rewrite, agent G): main-text notation (s_t, Loop(h_T), x, y^-, y^+), numbers as literals.
% Training-log endpoints (beta = 0.05, two runs that differ only in the likelihood anchor, alpha = 1):
%   DPO:        chosen -44.0 -> -51.1, rejected -40.6 -> -132.6  => 0.05*((-51.1+132.6)-(-3.4)) = 4.25
%   DPO+anchor: chosen -43.6 ->  -3.1, rejected -40.4 ->  -28.7  => 0.05*(( -3.1 +28.7)-(-3.2)) = 1.44
% Loop rates 6.1 / 27.0 / 37.7: tab:ablation (DPO, chosen-only SFT) and the CodeGym-only anchored row of
% tables/naacl_app/tab_registered_objective_controls.tex (all on the same 8,000 self-mined pairs).

\section{A Formal Account of Loops and Their Repair}
\label{app:theory}

\subsection{Setup and Definitions}
\label{app:theory:defs}

A trajectory $h_T=(a_1,o_1,\dots,a_T,o_T)$ alternates tool calls and
observations, and $h_{<t}$ is its prefix before step $t$; $s_t$ and
$\operatorname{Loop}(h_T)$ are as in \cref{sec:definition}. For a candidate
action $a$, $s(a;h_{<t})$ is the value $s_t$ would take if the agent issued
$a$ at step $t$. A context manager $C$ maps the history to the context the
policy acts on, $X_t=C(h_{<t})$, with $C=\mathrm{id}$ for the full history.
A self-mined pair (\cref{sec:diagnosis}) has prompt $x=X_t$, the context the
agent saw before the looping action $y^{-}=a_t$, and chosen action
$y^{+}=a_{t'}$.

\subsection{Loop Ambiguity and an Activity Ceiling}
\label{app:theory:ceiling}

Compression can map two histories that disagree on whether an action was
already taken to the same context, which then cannot tell a new action from a
looping one. Let \emph{work} $W_c$ count the actions other than
\texttt{Observe} and \texttt{Done} issued after the first compression event,
and define the \emph{loop ambiguity}
\[
\delta=\operatorname*{ess\,inf}\ \Pr\bigl[s(a;h_{<t})=1\,\big|\,X_t,\ a\text{ issued}\bigr]
\]
over these actions. We start after compression because the first action of
an episode never loops.

\paragraph{Proposition 1 (activity ceiling).}
A policy that cannot tell repeats from new actions loops less only by acting
less: if $\delta>0$ and the policy issues at most $\varepsilon$ looping actions
per task in expectation, then $\mathbb{E}[W_c]\le\varepsilon/\delta$.

\emph{Proof.} Each work action issued after compression loops with
conditional probability at least $\delta$, so
$\varepsilon\ge\mathbb{E}[\#\text{looping actions}]\ge\delta\,\mathbb{E}[W_c]$.
\hfill$\square$

Driving loops close to zero thus requires a lower $\delta$ or less activity, so
we report steps and tokens per task next to the loop rates of Table~\ref{tab:main}. Reading the action
record lowers $\delta$, yet the compressed baseline loops even with summaries
that record its past actions (\cref{sec:compression}); training the policy
targets this gap.

\subsection{Why Both Sides of a Pair Matter}
\label{app:theory:anchor}

The DPO gradient on a self-mined pair raises the loop-breaking action and
lowers the looping action under the same context. With
$z_\theta=\log\tfrac{\pi_\theta(y^{+}\mid x)}{\pi_{\mathrm{ref}}(y^{+}\mid x)}
-\log\tfrac{\pi_\theta(y^{-}\mid x)}{\pi_{\mathrm{ref}}(y^{-}\mid x)}$ and
$g=\sigma(-\beta z_\theta)$, the loss of \cref{sec:cure-method} is
$\mathcal{L}=-\log\sigma(\beta z_\theta)$ and
\begin{align*}
\nabla_\theta\mathcal{L}={}&-\beta g\,\nabla_\theta\log\pi_\theta(y^{+}\mid x)\\
&+\beta g\,\nabla_\theta\log\pi_\theta(y^{-}\mid x).
\end{align*}
Up to weights, the first term alone is chosen-only SFT, which does not break
the loop (\cref{tab:ablation}); the pair does.

\paragraph{A likelihood anchor dilutes the contrast.}
Adding $\alpha\,\ell_{\mathrm{nll}}$ with
$\ell_{\mathrm{nll}}=-\tfrac{1}{|y^{+}|}\log\pi_\theta(y^{+}\mid x)$ raises
the coefficient on the first term to $\beta g+\alpha/|y^{+}|$ and leaves the
second at $\beta g$. The \emph{contrast share}
\[
\rho(\alpha)=\frac{\beta g}{\beta g+\alpha/|y^{+}|}
\]
is 1 for DPO and falls toward 0, the chosen-only limit, as $\alpha$ grows,
so the loop rate should rise with $\alpha$.

On the 8,000 self-mined CodeGym pairs, the loop rate follows this order:
6.1\% for DPO (DeLoop), 27.0\% with the anchor at $\alpha=1$
(\cref{app:cure:rpo}), and 37.7\% for chosen-only SFT. In the training logs
of two runs that differ only in the anchor ($\beta=0.05$), DPO lowers the
loop-breaking action's mean log-probability only slightly, from $-44.0$ to
$-51.1$, and drives the looping action's from $-40.6$ to $-132.6$. With the
anchor, the loop-breaking action rises from $-43.6$ to $-3.1$ and the looping
action rises with it, from $-40.4$ to $-28.7$, so the implied reward margin
$\beta z_\theta$ ends at 1.44 instead of 4.25.

% NAACL/ARR appendix: extension studies (rewritten 2026-10-08 to match the main text).
% Negative results and caveats from the earlier version moved to the Limitations section.
\section{Extension Studies}
\label{app:ext}

We report four further studies on Qwen3-8B in CodeGym: what the policy
conditions on (\cref{app:ext:probe}), why the detector uses exact repetition
(\cref{app:ext:signatures}), a random-pair control
(\cref{app:ext:randompair}), and what the chosen action adds
(\cref{app:ext:chosen}).

\subsection{The Policy Reacts to the Record of Its Own Executed Action}
\label{app:ext:probe}

We probe the moments at which the agent loops. Unlike the training-side points
of \cref{app:probe}, these come from the compressed baseline's runs on the
300 evaluation tasks: 150 looping points, one per task at its first loop. We add 80 retry points, at which an earlier call to the same
action returned a different observation, so calling it again is reasonable.
At each point we rebuild the context from the raw history in six conditions:
\emph{not yet executed} (the history before the action was first run);
\emph{proposed only} (the same history plus a note that the action was
proposed but not run, padded to the same length); \emph{proposed +
observation} (the note also states what the action would return); \emph{just
executed} (the history ending right after the action's first run);
\emph{record present} (the full history up to the decision point); and
\emph{record deleted} (the same history with the record of the earlier run
removed). We sample eight completions per condition at temperature 0.7 and
report the re-issue rate, the share of completions whose next action is the
probed action (\cref{tab:ext-probe}).

\begin{table}[t]
\centering
\small
\setlength{\tabcolsep}{3pt}
\resizebox{\linewidth}{!}{% Re-issue rate (%) by context condition; Qwen3-8B CodeGym decision points, 8 samples at temperature 0.7.
% Values from extensions_20260908/_analysis/d1_probe/results_*_k8_t07*.json (base, our DPO policy, mixed-data policy).
\begin{tabular*}{\linewidth}{@{\extracolsep{\fill}}lrrrrrr@{}}
\toprule
\rowcolor{TableHead}
 & \multicolumn{3}{c}{Looping points ($n{=}150$)} & \multicolumn{3}{c}{Retry points ($n{=}80$)} \\
\rowcolor{TableHead}
Context condition & Base & DeLoop & DeLoop-Mix & Base & DeLoop & DeLoop-Mix \\ \midrule
Not yet executed & 73.7 & 61.4 & 57.3 & 68.4 & 63.3 & 62.2 \\
Proposed only & 73.9 & 41.0 & 24.4 & 66.7 & 31.9 & 25.2 \\
Proposed + observation & 29.0 & 2.5 & 2.2 & 20.6 & 4.7 & 7.2 \\
Just executed & 4.9 & 1.9 & 1.9 & 5.6 & 5.6 & 6.9 \\
Record present & 16.5 & 8.3 & 6.8 & 51.7 & 36.1 & 38.0 \\
Record deleted & 53.3 & 20.7 & 23.9 & 64.8 & 54.2 & 56.1 \\
\bottomrule
\end{tabular*}
}
\caption{\textbf{Action re-issue under different context conditions.}
Re-issue rate (\%) of the probed action under each context condition, at
looping points and at retry points in Qwen3-8B CodeGym runs. Base is the
untrained Qwen3-8B.}
\label{tab:ext-probe}
\end{table}

\paragraph{Untrained policy.} Planning an action does not stop the agent from
issuing it: with the action only proposed, the untrained Qwen3-8B re-issues it
in 73.9\% of completions, the same as before the action was ever run
(73.7\%). Stating the observation lowers re-issue to 29.0\%, and the record of
the agent's own execution lowers it further, to 16.5\%. Deleting that record
brings re-issue back to 53.3\%. The effect is strongest right after the run
(4.9\%) and fades as history accumulates.

The record also separates loops from reasonable retries. At retry points it
lowers re-issue only from 68.4\% to 51.7\%, while a stated observation alone
lowers it to 20.6\% and blocks retries that are worth making. The executed
record, together with what it returned, is what tells the policy whether an
action is finished or worth checking again.

\paragraph{Trained policies.} Both trained policies still tell loops from
retries. With the record present, DeLoop re-issues a looping action in
8.3\% of completions and DeLoop-Mix in 6.8\%, against 36.1\% and
38.0\% at retry points. Deleting the record raises re-issue at looping points
only to 20.7\% and 23.9\%, against 53.3\% for the untrained model, while
retries return to 54.2\% and 56.1\%. The trained policies thus avoid most
loops even when the record of the earlier run is missing.

\subsection{Compression Amplifies Exact Repetition Most}
\label{app:ext:signatures}

Our detector flags only exact repeats with an identical observation. We
measure four broader signatures on the same CodeGym runs:
\emph{near-duplicate} calls (same tool, different arguments, with an
identical observation or a small argument edit),
\emph{argument thrash} (A--B--A argument alternation within three calls),
\emph{redundant observe} (an action whose observation is already in context,
word for word, from a different action), and \emph{productive repeats} (the
same action with a different observation). \Cref{tab:ext-signatures} gives
the results.

\begin{table*}[t]
\centering
\small
\setlength{\tabcolsep}{4pt}
% Repetition signatures on Qwen3-8B CodeGym evaluation runs (300 tasks; pooled over 3 runs; DeLoop = the three runs behind its 25.9% success, recomputed 2026-10-11).
% Values from extensions_20260908/_analysis/exp01_signatures.json (prevalence, per_step_rates_pooled,
% compression_sensitivity_summ_over_none.per_step_rate_ratio, base_none association odds_ratio).
\begin{tabular*}{\linewidth}{@{\extracolsep{\fill}}lrrrrrrrr@{}}
\toprule
\rowcolor{TableHead}
 & \multicolumn{2}{c}{Tasks (\%)} & \multicolumn{2}{c}{Per action (\%)} & Ratio & \multicolumn{2}{c}{Trained, per action (\%)} & Failure \\
\rowcolor{TableHead}
Signature & Full & Comp. & Full & Comp. & ($\times$) & DeLoop & DeLoop-Mix & OR \\ \midrule
\rowcolor{TableOurs} Exact loop (our detector) & 28.6 & 57.0 & 6.6 & 24.8 & 3.77 & 2.3 & 2.5 & 1.62 \\
Near-duplicate & 75.7 & 77.9 & 46.7 & 34.3 & 0.73 & 19.0 & 23.7 & 0.68 \\
Argument thrash & 3.3 & 3.4 & 0.4 & 0.3 & 0.92 & 0.1 & 0.2 & 3.10 \\
Redundant observe & 52.3 & 56.3 & 21.1 & 22.5 & 1.07 & 6.7 & 7.7 & 0.69 \\
Productive repeat & 30.0 & 39.2 & 8.9 & 10.7 & 1.21 & 5.9 & 7.6 & 2.00 \\
\bottomrule
\end{tabular*}

\caption{\textbf{Repetition signatures under compression.} Qwen3-8B on the 300 CodeGym evaluation tasks. Tasks: share of
tasks with at least one event; per action: share of actions that are events,
with the full history (Full) and under summarization at 4,096 tokens (Comp.);
ratio: Comp./Full per action. The DeLoop and DeLoop-Mix columns are
under summarization. Failure OR: odds ratio of
task failure for tasks with the signature, with the full history. Rates are
recomputed from the logged trajectories of three evaluation runs, and the
exact-loop task shares differ slightly from the run averages in Table~\ref{tab:main}.}
\label{tab:ext-signatures}
\end{table*}

Exact loops are the signature that compression amplifies most. Under
summarization, their share of actions rises 3.77$\times$, from 6.6\% to
24.8\%, while every broader signature changes by 0.73$\times$ to
1.21$\times$. Near-duplicate and redundant-observe calls are common, in
77.9\% and 56.3\% of compressed tasks, but they behave like ordinary
re-probing: with the full history they are more frequent in solved tasks
(failure odds ratios 0.68 and 0.69). We therefore define the loop as exact
repetition. Training lowers every signature per action, so the trained
policies do not trade exact loops for near-duplicates: near-duplicate calls
fall from 34.3\% of actions to 19.0\% under DeLoop and 23.7\% under
DeLoop-Mix.

\subsection{Random Pairs Do Not Break the Loop}
\label{app:ext:randompair}

Does the loop label matter, or would any preference pair from the agent's
own rollouts do? We build a random-pair corpus from the compressed baseline's
CodeGym rollouts: each pair keeps a prompt and takes as its chosen and
rejected actions two random later actions from the same rollout. We train on
877 random pairs and on random subsets of our 8,000 self-mined pairs with the
same DPO recipe, match the number of optimizer updates to our main recipe
(about 500), and evaluate on a separate 128-task held-out CodeGym set
(\cref{tab:ext-d7c}).

\begin{table}[t]
\centering
\small
\setlength{\tabcolsep}{3pt}
\resizebox{\linewidth}{!}{% Random-pair control at the main recipe's update budget; Qwen3-8B, 128-task held-out CodeGym set, summarization at 4,096 tokens.
% Values from extensions_20260908/_analysis/exp09_d7b.json (d7c arms) and exp04_d7.json (compressed baseline, pooled over 3 runs).
\begin{tabular*}{\linewidth}{@{\extracolsep{\fill}}llrrrr@{}}
\toprule
\rowcolor{TableHead}
Training data & Pairs & Updates & Loop (\%) $\downarrow$ & Success (\%) $\uparrow$ & Steps $\downarrow$ \\ \midrule
\rowcolor{TableBase} Compressed baseline & --- & --- & 66.1 & 11.5 & 19.12 \\
Random pairs & 877$\times$1 & $\approx$55 & 54.8 & 11.9 & 13.26 \\
Random pairs & 877$\times$9 & $\approx$495 & 69.5 & 11.7 & 16.17 \\
\rowcolor{TableOurs} Self-mined pairs & 1,003$\times$8 & $\approx$504 & 9.4 & 23.4 & 5.82 \\
\rowcolor{TableOurs} Self-mined pairs & 4,012$\times$2 & $\approx$502 & 5.5 & 20.3 & 4.95 \\
\rowcolor{TableOurs} Self-mined pairs (DeLoop) & 8,000$\times$1 & $\approx$500 & 5.5 & 15.6 & 4.80 \\
\bottomrule
\end{tabular*}
}
\caption{\textbf{Random-pair control.}
Qwen3-8B on another 128-task held-out CodeGym set under summarization at 4,096
tokens (Following setting from~\cite{acon}. Each row trains with the same DPO recipe; Pairs gives pairs
$\times$ epochs and Updates the number of optimizer updates. The compressed
baseline is pooled over three evaluation runs.}
\label{tab:ext-d7c}
\end{table}

At the same budget of about 500 optimizer updates, 1,003 self-mined pairs bring
the loop rate to 9.4\% and task success to 23.4\%, while random pairs leave
the loop rate at 69.5\%, above the compressed baseline's 66.1\%, and task
success at 11.7\%. Training longer on random
pairs makes looping worse, from 54.8\% after one epoch to 69.5\% after nine.
With 4,012 self-mined pairs the loop rate reaches 5.5\%, the same as
DeLoop trained on all 8,000. What breaks the loop is the loop label, not
preference training on the agent's own rollouts.

\subsection{The Loop-Breaking Action Matters}
\label{app:ext:chosen}

Each self-mined pair prefers the loop-breaking action to the looping action.
To test what the chosen side adds, we keep the same 8,000 prompts, the same
rejected actions, and the same recipe, and replace the chosen action in two
ways: with a random action from the prompt's own history, or with the chosen
action of a random other pair, a plausible action from the wrong state
(\cref{tab:ext-e1}).

\begin{table}[t]
\centering
\small
\setlength{\tabcolsep}{3pt}
\resizebox{\linewidth}{!}{% Chosen-action replacement (success column removed in review 2026-10-08); Qwen3-8B, 300 CodeGym evaluation tasks, summarization at 4,096 tokens.
% Values from extensions_20260908/_analysis/exp07_e1.json (every row pooled over 3 runs; DeLoop = the three runs behind its 25.9% success, recomputed 2026-10-11).
\begin{tabular*}{\linewidth}{@{\extracolsep{\fill}}lrr@{}}
\toprule
\rowcolor{TableHead}
Chosen action & Loop (\%) $\downarrow$ & Steps $\downarrow$ \\ \midrule
\rowcolor{TableBase} Compressed baseline (no training) & 57.0 & 14.91 \\
\rowcolor{TableOurs} Loop-breaking action (DeLoop) & 5.7 & 4.59 \\
Random action from the same history & 41.9 & 8.17 \\
Chosen action of another pair & 36.1 & 8.99 \\
\bottomrule
\end{tabular*}
}
\caption{\textbf{Chosen-action controls.} Qwen3-8B
on the 300 CodeGym evaluation tasks under summarization at 4,096 tokens. All
trained rows use the same prompts, rejected actions, and DPO recipe and differ
only in the chosen action. Rows are recomputed from the logged trajectories of three evaluation runs and
differ slightly from the run averages in
Table~\ref{tab:main}.}
\label{tab:ext-e1}
\end{table}

The loop-breaking action carries more than ``anything but the looping
action''. With a random chosen action, the loop rate stays at 41.9\% (random
action from the history) and 36.1\% (chosen action of another pair), against
5.7\% with the loop-breaking action. Together with \cref{tab:ablation}, this
shows that both sides of each pair carry the signal.

% NAACL appendix: four further analyses of existing runs (added 2026-10-12, rewritten the same day).
% Sources (CPU re-analyses of archived runs, no new rollouts), all scored with the Table 1 scorer:
%   Anti_loop/extensions_20260908/_analysis/exp19_paperscorer_checks.{md,json} (ext_analysis_19_paperscorer_checks.py):
%     loop split by whether the repeated call is still in context (context replayed from the logged
%     compression events, trusting each logged compressed_turns); Gemini same-task / length-matched / all-task rates.
%   Anti_loop/extensions_20260908/_analysis/exp18_blocker_gemini.md (ext_analysis_18_blocker_gemini.py): blocker rows.
%   J.2 tokens/task: tools/efficiency_metrics.py on guard_dupblock_test300d (24.9k) and nf8k_dupblock_test300d_summ (12.8k).
%   Anti_loop/extensions_20260908/_analysis/exp20_termination.{md,json} (ext_analysis_20_termination.py): Sec. J.4.
% DeLoop = the three runs behind Table 1 (nf8k_r2_test300d_summ, nf8k_test300d_rep3, nf8k_test300d_rep5).
\section{Additional Analyses}
\label{app:checks}

This section reports four further analyses of the runs behind
Table~\ref{tab:main} and Figure~\ref{fig:compression}. None of them needs new
rollouts.

\subsection{Loops on Calls Still in Context}
\label{app:checks:context}

\begin{coltable}[t]
\centering
\small
\setlength{\tabcolsep}{4pt}
\begin{tabular*}{\linewidth}{@{\extracolsep{\fill}}lccc@{}}
\toprule
\rowcolor{TableHead}
Policy & Loop rate & In context & Summarized \\ \midrule
\rowcolor{TableBase} Compressed baseline & 57.3 & 25.2 & 32.1 \\
\rowcolor{TableOurs} DeLoop & 6.1 & 3.2 & 2.9 \\
DeLoop-Mix & 10.1 & 4.3 & 5.8 \\
Full history & 29.0 & 29.0 & -- \\
\bottomrule
\end{tabular*}
\caption{\textbf{Loops by availability of the earlier record.} Loop rate (\%) of Qwen3-8B on CodeGym under summarization at 4,096
tokens, averaged over three runs; the full-history agent uses no compression.
A run counts under \emph{In context} if it repeats a call that is still in its
context, and under \emph{Summarized} if every call it repeats had been
summarized.}
\label{tab:checks-context}
\end{coltable}

Our loop rate compares each call with every earlier call in the run. Under
compression, some of those earlier calls are no longer in the agent's context
word for word, because the harness has replaced them with a summary. We
therefore also count a loop only when the earlier call and its observation are
still in the context the agent sees. We rebuild this context at every step from
the compression events in the logs (\cref{app:setup}).

DeLoop removes loops whether or not the earlier call is still in context
(\cref{tab:checks-context}). Counting only repeats of calls still in context,
DeLoop lowers the loop rate of Qwen3-8B on CodeGym from 25.2\% to 3.2\%. For
repeats of calls that had been summarized, it lowers the rate from 32.1\% to
2.9\%.

The split also shows where compression adds loops. The compressed agent repeats
a call that is still in its context in 25.2\% of runs, close to the 29.0\% loop
rate of the full-history agent. The added loops come from calls that the harness
has summarized, even though the summarization prompt asks the model to keep
executed actions and their results (\cref{app:setup}).

\subsection{The Duplicate Blocker}
\label{app:checks:blocker}

\begin{coltable}[t]
\centering
\small
\setlength{\tabcolsep}{3pt}
\resizebox{\linewidth}{!}{%
\begin{tabular}{@{}lccccc@{}}
\toprule
\rowcolor{TableHead}
 & \multicolumn{2}{c}{Repeat proposals} & Loop & Steps & Success \\ \cmidrule(lr){2-3}
\rowcolor{TableHead}
Policy & Tasks (\%) & Per task & (\%) & & (\%) \\ \midrule
\rowcolor{TableBase} Compressed baseline & 69.8 & 5.29 & 57.3 & 15.00 & 25.1 \\
Duplicate blocker & 71.3 & 4.77 & 5.7 & 14.71 & 25.3 \\
\rowcolor{TableOurs} DeLoop & 13.6 & 0.38 & 6.1 & 4.67 & 25.9 \\
DeLoop + blocker & 13.7 & 0.25 & 1.3 & 4.84 & 27.7 \\
\bottomrule
\end{tabular}}
\caption{\textbf{Proposed repeats and executed loops with a duplicate blocker.} Qwen3-8B on CodeGym under summarization at 4,096 tokens;
three runs for the compressed baseline and DeLoop, one run for each blocker row.
\emph{Repeat proposals}: tasks in which the agent proposes a call it has already
run, and such proposals per task. \emph{Loop}: loops among the calls that run
(\cref{sec:definition}).}
\label{tab:checks-blocker}
\end{coltable}

The exact duplicate blocker refuses to run a call that repeats an earlier one
and returns the earlier result with a warning instead (\cref{app:cure:inference}).
Its loop rate in Table~\ref{tab:inference-controls}, 46.7\%, counts these
refused calls, since they never run. Counting only the calls that run, as for
the other rows, its loop rate is 5.7\%. Neither number shows what the agent
does, so we also count how often it proposes a call it has already run
(\cref{tab:checks-blocker}).

The blocker does not change what the agent proposes. With the blocker,
Qwen3-8B proposes an already executed call in 71.3\% of tasks, close to the
69.8\% without it, and steps and success barely move. The agent still
generates every refused call in full, so the blocker saves no model compute: it uses 24.9k tokens per task, against 23.9k without it and
12.7k for DeLoop (Table~\ref{tab:inference-controls}). DeLoop lowers repeat
proposals to 13.6\% of tasks and cuts steps per task from 15.00 to 4.67.

The two methods are complementary. The blocker stops the few repeats that
DeLoop still proposes, and DeLoop removes most of the calls that the blocker
would refuse only after they are generated. With the blocker on, DeLoop runs a
repeated call in 1.3\% of tasks, uses 12.8k tokens per task, and solves 27.7\%
of them in a single run.

\subsection{Gemini Models on the Same Tasks}
\label{app:checks:gemini}

\begin{coltable}[t]
\centering
\small
\setlength{\tabcolsep}{2pt}
\begin{tabular*}{\linewidth}{@{\extracolsep{\fill}}lccccc@{}}
\toprule
\rowcolor{TableHead}
 & & \multicolumn{2}{c}{Same tasks} & \multicolumn{2}{c}{All tasks} \\ \cmidrule(lr){3-4}\cmidrule(lr){5-6}
\rowcolor{TableHead}
Model & Tasks & None & Summ. & None & Summ. \\ \midrule
Gemini 3.5 Flash-Lite & 191 & 40.8 & 47.1 & 34.0 & 36.0 \\
Gemini 3.1 Flash & 226 & 46.5 & 65.5 & 40.0 & 52.7 \\
Gemini 3.1 Pro & 139 & 33.1 & 61.9 & 21.3 & 34.3 \\
\bottomrule
\end{tabular*}
\caption{\textbf{Gemini loop rates on matched task sets.} Loop
rate (\%) on CodeGym without compression (None) and with summarization at 1,400
tokens (Summ.). \emph{Same tasks}: the tasks on which compression fired in the
summarization run. One run per model and setting.}
\label{tab:checks-gemini}
\end{coltable}

Figure~\ref{fig:compression}b compares each Gemini model's full-history run
with the episodes of its summarization run in which compression fired.
Compression fires only in longer episodes, and longer episodes loop more often
even without compression. We therefore also compare the two runs on the same
tasks, the ones on which compression fired (\cref{tab:checks-gemini}).

Compression raises the loop rate of every model on the same tasks. Gemini 3.1
Pro loops in 61.9\% of these tasks with compression and in 33.1\% without it,
1.9$\times$ as often (2.9$\times$ in Figure~\ref{fig:compression}b). The ratio
is 1.4$\times$ for Gemini 3.1 Flash and 1.2$\times$ for Gemini 3.5 Flash-Lite.
Comparing against full-history episodes long enough for compression to fire
gives similar ratios (1.8$\times$, 1.4$\times$, and 1.1$\times$). Over all 300
tasks, including short episodes in which compression never fires, the ratios
are 1.6$\times$, 1.3$\times$, and 1.1$\times$.

\subsection{Fewer Loops Are Not Just Shorter Episodes}
\label{app:checks:length}

\begin{coltable}[t]
\centering
\small
\setlength{\tabcolsep}{4pt}
\begin{tabular*}{\linewidth}{@{\extracolsep{\fill}}lcccc@{}}
\toprule
\rowcolor{TableHead}
 & \multicolumn{4}{c}{Step index} \\ \cmidrule(lr){2-5}
\rowcolor{TableHead}
Policy & 0--4 & 5--9 & 10--14 & 15--19 \\ \midrule
\rowcolor{TableBase} Compressed baseline & 4.1 & 22.3 & 32.8 & 42.8 \\
\rowcolor{TableOurs} DeLoop & 0.8 & 4.1 & 7.0 & 25.7 \\
DeLoop-Mix & 1.8 & 3.6 & 4.5 & 7.3 \\
Full history & 2.3 & 6.8 & 11.3 & 13.9 \\
\bottomrule
\end{tabular*}
\caption{\textbf{Loop hazard by step index.} Loop hazard
(\%): the share of steps that are looping actions among the steps an episode
reaches, by step index. Qwen3-8B on CodeGym under summarization at 4,096 tokens
(full history: no compression), pooled over three runs per policy.}
\label{tab:checks-hazard}
\end{coltable}

DeLoop shortens episodes from 15.00 to 4.67 steps (Table~\ref{tab:main}), and a
shorter episode has fewer chances to loop. We therefore measure looping in two
ways that do not depend on episode length.

First, we compute the loop hazard: among the steps that episodes reach at a
given step index, the share that are looping actions
(\cref{tab:checks-hazard}). DeLoop lowers it at every step index, from 22.3\%
to 4.1\% at steps 5--9 and from 32.8\% to 7.0\% at steps 10--14. Second, we
compare episodes of similar length. Among episodes of 6--10 steps, the loop rate
falls from 33.2\% to 13.1\%, and among episodes of 11--20 steps from 69.1\% to
29.2\%.

DeLoop also answers earlier without losing accuracy on those tasks. It answers
within four steps on average on 164 of the 300 tasks. On these tasks it solves
30.1\% of runs in 2.7 steps, and the compressed baseline solves the same 30.1\%
in 13.0 steps (paired difference $+0.0$ points, 95\% bootstrap interval over
tasks $[-6.3, +6.1]$). The result is similar with a cutoff of three to six
steps.
\FloatBarrier

\end{document}